\documentclass[letterpaper,twocolumn,11pt]{article}
\usepackage{usenix}

\usepackage{tikz}
\usepackage{amsmath}
\usepackage{amsmath}
\usepackage{amssymb}
\usepackage{booktabs}
\usepackage{multirow}
\usepackage{diagbox}
\usepackage{url}
\usepackage{enumitem}
\usepackage{makecell}
\usepackage{tabularx} 
\usepackage{subcaption}
\newtheorem{theorem}{Theorem}

\newtheorem{definition}{Definition}

\usepackage{filecontents}

\DeclareMathOperator*{\argmin}{argmin}
\usepackage{graphicx} 
\usepackage{subcaption} 
\usepackage{booktabs}
\usepackage{colortbl}
\usepackage{bm}
\usepackage{makecell}

\definecolor{mygray}{gray}{.9}
\definecolor{mypink}{RGB}{254, 241, 240}
\definecolor{myyellow}{RGB}{255,242,204}
\definecolor{mylightgreen}{RGB}{235, 251, 251}

\usepackage{changes} 
\definecolor{addcolor}{rgb}{0.85, 0.93, 0.82}
\definecolor{modcolor}{rgb}{1.0, 0.92, 0.8}  

\usepackage{xspace}
\xspaceaddexceptions{[]\{\}}
\newcommand{\InvLoss}{{\sf InvLoss}\xspace}
\newcommand{\InvL}{{\sf InvL}\xspace}
\newcommand{\InvRE}{{\sf InvRE}\xspace}

\usepackage[normalem]{ulem}

\newcommand*{\qedb}{\hfill\ensuremath{\square}}   

\begin{document}

\date{}

\title{From Risk to Resilience: Towards Assessing and Mitigating the Risk of Data Reconstruction Attacks in Federated Learning}

\author{
{\rm Xiangrui Xu\thanks{The work of Xiangrui Xu was conducted during her time as a Visiting Research Student in Zhize Li's group at SMU.}}\\
Beijing Jiaotong University \& \\
Singapore Management University
\and
{\rm Zhize Li}\\
Singapore Management University
\and
{\rm Yufei Han}\\
INRIA Rennes-Bretagne-Atlantique
\and
{\rm Bin Wang}\\
Zhejiang Key Laboratory of AIoT Network and Data Security
\and
{\rm Jiqiang Liu}\\
Beijing Jiaotong University
\and
{\rm Wei Wang\thanks{Corresponding author (\texttt{wangwei1@bjtu.edu.cn}).}}\\
Beijing Jiaotong University \& 
Xi'an Jiaotong University
} 

\maketitle

\begin{abstract}

Data Reconstruction Attacks (DRA) pose a significant threat to Federated Learning (FL) systems by enabling adversaries to infer sensitive training data from local clients. Despite extensive research, the question of how to characterize and assess the risk of DRAs in FL systems remains unresolved due to the lack of a theoretically-grounded risk quantification framework.  In this work, we address this gap by introducing \emph{Invertibility Loss} (\InvLoss) to quantify the maximum achievable effectiveness of DRAs for a given data instance and FL model. We derive a tight and computable upper bound for \InvLoss and explore its implications from three perspectives. First, we show that DRA risk is governed by the spectral properties of the Jacobian matrix of exchanged model updates or feature embeddings, providing a unified explanation for the effectiveness of defense methods. Second, we develop \InvRE, an \InvLoss-based DRA risk estimator that offers attack method-agnostic, comprehensive risk evaluation across data instances and model architectures. Third, we propose two adaptive noise perturbation defenses that enhance FL privacy without harming classification accuracy. Extensive experiments on real-world datasets validate our framework, demonstrating its potential for systematic DRA risk evaluation and mitigation in FL systems.

\end{abstract}

\section{Introduction} \label{sec:intro}

Federated Learning (FL) has emerged as a promising paradigm to address data silo challenges and comply with privacy regulations such as GDPR~\cite{first_paper, FL_survey1, FL_survey4}. FL can be categorized into Horizontal FL (HFL) and Vertical FL (VFL) based on partitioning of the sample and feature spaces~\cite{FL_survey1, FL_survey3, coresetVFL}.  
In HFL, clients share a common feature extractor while working on different subsets of training data, making it ideal for collaborative tasks like personalized recommendations (e.g., enhancing a global next-word prediction model using keystroke data from Google users~\cite{FL_survey2}). In contrast, VFL involves a shared subject set with partitioned feature spaces, commonly applied in fields like finance and healthcare. Entities such as banks or hospitals contribute distinct private features for the same subjects to enable predictive analysis~\cite{FL_survey4}.

Recent studies~\cite{FL_survey2, FL_survey4} reveal that Federated Learning systems, including HFL and VFL, are vulnerable to privacy-stealing attacks. In HFL, adversaries at the central server can access model parameters and updates from participants, enabling membership inference attacks (MIA)~\cite{Mem1, Mem2} or data reconstruction attacks (DRA) to infer training data~\cite{DLG, iDLG, IG, GI, GGL, CGIR}. In VFL, privacy risks are heightened during prediction. Participants submit feature embeddings derived from their private data to a top-layer server in a split-learning setup~\cite{VFL_ICDE, PriVFL}. Adversaries controlling the server can exploit these embeddings to reconstruct private feature values, violating data privacy~\cite{Ressfl, VFL_cafe, PISTE, VFL_label, VFL_unleashing, VFLMonitor}. In this work, we focus on DRA attacks, a critical privacy vulnerability in FL. Existing defenses either lack strong theoretical guarantees~\cite{FL_eval}, or fail to provide consistent protection across FL tasks without significantly compromising utility~\cite{DLG, IG}. These limitations make DRA attacks a critical privacy vulnerability in FL.

Defending DRA attacks in FL involves two critical aspects. \textbf{First}, accurately assessing the feasibility and effectiveness of DRA attacks is essential. The central question is: \textit{Given an FL application defined by its model architecture and data distribution, how effective are DRA attacks at achieving high data reconstruction accuracy?} The model updates or embeddings shared by FL clients can be viewed as a transformation of the input data. DRA attacks aim to invert this transformation~\cite{DLG, PISTE}, with their effectiveness determined by the computational feasibility of finding this inverse, also known as the transformation’s invertibility. However, the highly non-linear nature of deep neural networks (DNNs) in FL models makes it challenging to theoretically and empirically quantify this invertibility and its associated privacy risks. 
\textbf{Second}, FL stakeholders must select model architectures that enhance privacy protection and understand how privacy risks vary across data sources~\cite{FL_eval}. Accurate DRA risk estimation is crucial for measuring and comparing risks across different input samples and models, enabling informed decisions to strengthen privacy in FL systems.

Most prior studies to assess DRA risks~\cite{defense_pruning, defense_dropout, Ressfl, NoPeek} use an \emph{assess-by-attack} strategy, evaluating privacy protection by conducting specific DRAs and analyzing reconstruction accuracy. Higher accuracy indicates more effective attacks and greater vulnerability of FL systems. However, this strategy is inherently unreliable, as it risks underestimating vulnerabilities. Advanced, carefully crafted attacks could still succeed where earlier, simpler attempts failed. Recent efforts explore information-theoretic metrics to quantify DRA risks in FL, such as mutual information (MI)-based~\cite{MI-estimator, MI-estimator2, PriVFL} and Fisher information-based approaches (dFIL)~\cite{Fisherinfo_theory2, VFL_Fisherinfo}. For instance, Tan et al.~\cite{MI-estimator} proposed using MI between data and model parameters to assess the feasibility of DRA attacks in HFL. However, directly computing the MI metric for non-linear FL models is non-trivial, leading to approximations that focus on data-space MI. These approximations fail to capture correlations between training data and model parameters, which limits their accuracy in evaluating privacy risks across diverse model architectures. Similarly, dFIL~\cite{VFL_Fisherinfo} uses Fisher information to establish a lower bound on reconstruction error. However, the applicability of dFIL \cite{VFL_Fisherinfo} is constrained by its dependence on an unbiased data reconstruction estimator, whereas most DRA attacks rely on biased estimators. Moreover, dFIL cannot effectively compare DRA risks across different model architectures. In addition, both methods are tailored to specific FL mechanisms (HFL or VFL), leaving the intrinsic connections between DRA vulnerabilities across mechanisms unexplored. These limitations highlight the need for systematic assessment and comparison of DRA risks across various FL mechanisms, datasets, and model architectures.

Our work addresses the key limitations in DRA attack analysis by characterizing and assessing the invertibility of the transformation from data instances to model space. We first introduce \textit{Invertibility Loss} (\InvLoss), which quantifies the feasibility of DRAs by measuring the minimum reconstruction error achievable through an inverse transformation that maps model updates or feature embeddings exchanged in FL back to data space. Lower \InvLoss values indicate a higher DRA risk. We unveil that DRAs minimize \InvLoss, which is in nature equivalent to approximating the Moore-Penrose inverse of the Jacobian matrix of the observed model updates or embeddings (accessible to the attacker) with respect to input data. The upper bound of \InvLoss in FL is thus determined by the spectral properties of the Jacobian matrix, which can be efficiently computed using Singular Value Decomposition (SVD). Unlike previous approaches, our bound does not assume unbiased DRA estimators, offering a more accurate quantification of information leak risks. This enables a unified framework to assess DRA risks across both HFL and VFL, and compare risks across data instances and model architectures.

We further extend the \InvLoss framework to understand how the existing noise perturbation and information compression methods increase reconstruction error and reduce DRA risk. We propose two adaptive noise perturbation strategies that optimize noise injection based on the spectrum of the Jacobian matrix, providing stronger privacy protection without sacrificing classification accuracy. Finally, we introduce \InvRE, an exact estimator of \InvLoss, validated through extensive empirical evaluation as a reliable indicator of DRA risk across diverse models and data sources.

\textbf{Contributions.} We summarize our contribution in this study from three perspectives:
\begin{itemize}

\item We introduce Invertibility Loss (\InvLoss) as a feasibility measurement of DRA attacks in FL. To evaluate the DRA risk, we establish the upper of \InvLoss, forming a unified theoretical reasoning about the main factors determining the effectiveness of DRA attacks in different FL mechanisms. We further extend the theoretical analysis to explain the effectiveness of the existing defense mechanisms against DRA attacks.

\item We develop \InvRE for practical DRA risk estimation based on the theoretical upper bound of \InvLoss. \InvRE quantifies and compares DRA attack feasibility across various FL mechanisms, model architectures, and data sources, independent of specific attack methods. We validate \InvRE using 4 diverse datasets, 3 widely-used FL model architectures, and 3 state-of-the-art DRA attack methods for both HFL and VFL. Results confirm that \InvRE consistently reflects DRA risk levels across different tasks with/without privacy defense methods.

\item Leveraging the insights from the theoretical analysis, we propose two adaptive noise perturbation strategies that dynamically modulate noise injection based on the spectral properties of the Jacobian matrix. Empirical results confirm that they can reach strong privacy protection without harming the utility of FL systems.

\end{itemize}

\section{Preliminaries and Related Work} \label{sec:preliminaries}

Let $\bm{x} \in \mathbb{R}^{m}$ represent a data instance hosted by the client in FL (HFL or VFL) systems.
$\mathcal{F}(\bm{x}) \in \mathbb{R}^{p}$ denotes the shared model gradients or feature
embedding vectors in HFL or VFL respectively. 
Table~\ref{table:Notations} summarizes the notations in our work.

\noindent \textbf{Federated Learning.}
FL enables a group of remote clients $U = \{ u_1, \cdots , u_n \}$ to collaboratively train a machine learning model. Each client $u_i$ contributes a private set of training data, which remains inaccessible to other participants due to privacy concerns.
Based on the partitioning of sample and feature spaces, FL is classified into Horizontal FL (HFL) and Vertical FL (VFL)~\cite{FL_survey2, FL_survey3}.  

\begin{table}[t]
\centering
\caption{Notations and description.} 
\renewcommand\arraystretch{1} 
\resizebox{0.98\linewidth}{!}{
\begin{tabular}{c|c}
\toprule
\textbf{Notations}        & \textbf{Description}                   \\ \midrule
$\bm{x} \in \mathbb{R}^{m}$        & Input data from the client      \\  
$\mathcal{F}(\cdot)$      & Transformation function applied to $\bm{x}$: $\mathbb{R}^m\mapsto \mathbb{R}^p$  \\ 
$\mathcal{A}_{\bm{x}}(\cdot)$      & The inverse function applied to $\mathcal{F}(\cdot)$: $\mathbb{R}^p\mapsto \mathbb{R}^m$ \\  
$\bm{G_{x}}$                  
& Jacobian matrix $\partial \mathcal{F}(\bm{x})/\partial\bm{x}$ w.r.t a data instance\\ 
$\bm{A_{x}}$                  & The inverse matrix for data reconstruction  \\
$\bm{U}$,  $\bm{\Sigma}$,  $\bm{V}$   & \makecell{Left-singular vectors, matrix of singular \\ values, and right-singular vectors of $\bm{G_{x}}$}\\
$\sigma_{i}$                  & $i$-th singular value of the Jacobian matrix $\bm{G_{x}}$   \\ 
$\bm{\epsilon}$           & Noise matrix\\
$\bm{n}$                  &  Noise matrix in the singular space\\ 
\bottomrule
\end{tabular}
\vspace{-3mm}
\label{table:Notations}
}
\end{table}
\begin{figure*}[htbp]
    \centering
    \begin{subfigure}[b]{0.475\textwidth}
        \centering
        \includegraphics[width=\textwidth]{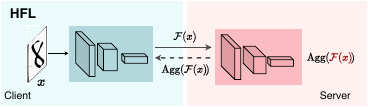} %
        \caption{HFL}
        \label{fig:hfl}
    \end{subfigure}
    \hfill
    \begin{subfigure}[b]{0.475\textwidth}
        \centering
        \includegraphics[width=\textwidth]{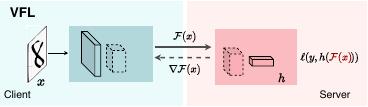} 
        \caption{VFL}
        \label{fig:vfl}
    \end{subfigure}
    \caption{Federated Learning architecture.}
    \vspace{-3mm}
    \label{fig:FL_workflow}
\end{figure*}

In HFL, clients collaboratively train a global model by sharing model updates with a central server, as shown in Fig.~\ref{fig:hfl}. At each iteration \(t\), the server selects a subset of \(n\) clients and shares the global model \(\bm{\theta}^t\) with them.  Each client $u_{i}$ then perform local training to compute local model updates \(\Delta {\bm{\theta}^{t}_i}\). These model updates are sent to the server for aggregation, typically using a weighted average: 
\(\bm{\theta}^{t+1} = \text{Agg}(\Delta{\bm{\theta}_1^{t}}, \dots, \Delta{\bm{\theta}_n^{t}}; \bm{\theta}^t)\), producing the updated model \(\bm{\theta}^{t+1}\). This process repeats until convergence, after which clients use the trained global model for local predictions. 
VFL uses a split-learning framework for both training and inference, as shown in Fig.~\ref{fig:vfl}~\cite{Splitfed, Ressfl, VFL_unleashing}. Clients host the bottom model for feature embedding generation. The server manages the top model to aggregate the local clients' feature embeddings for classification. During training, intermediate feature embeddings and gradients of bottom models are exchanged between the client and server until a collaboratively trained model is obtained. After training, bottom models can be transferred to the server for evaluation. In prediction, the server sends the input sample's ID to the clients. All clients then collaboratively compute the final classification output.

\noindent \textbf{Data Reconstruction Attacks in FL.} 
In data reconstruction attack scenarios involving FL systems, the attacker can either observe the model updates submitted by local clients in HFL~\cite{DLG, IG, CGIR} or analyze the embedding vectors generated by the bottom models owned by local clients in VFL~\cite{VFL_cafe, PISTE, VFL_label}. 
The model updates or the embeddings can be considered as a functional transformation $\mathcal{F}$ of the input instance $\bm{x}$ held by the clients, i.e., $\mathcal{F}(\bm{x})$.
The goal of the attacker can be thus defined as a function-matching problem~\cite{DLG}: 

\begin{equation}\label{eq:dra_obj}
\hat{\bm{x}}^{*} = \underset{\hat{\bm{x}}}{\argmin}\| \mathcal{F}(\bm{x}) - \mathcal{F}(\hat{\bm{x}})\|,
\end{equation}
where $\hat{\bm{x}}$ represents an estimate of the true input $\bm{x}$, and $\parallel \cdot \parallel$ denotes the Euclidean distance. Minimizing Eq.~\eqref{eq:dra_obj} aims to identify the reconstructed instance $\hat{\bm{x}}$ that most closely approximates the functional output of the true input $\bm{x}$.
This approach was initially proposed in the context of HFL with gradient matching~\cite{DLG}, it has been extended to VFL via embedding matching, as shown in~\cite{VFL_cafe}. Subsequent research has demonstrated that incorporating prior knowledge about the input can enhance this regression task~\cite{IG, GI, GGL, mGAN-AI}.
\emph{Inverting Gradients (IG)}~\cite{IG} proposed to employ regularization techniques, such as the total variation (TV) prior, to mitigate high-frequency artifacts in $\hat{\bm{x}}$ with cosine distance as the distance metric. 
Similar approaches can also be adapted for VFL, as detailed in~\cite{VFL_cafe}. More recent studies~\cite{CGIR, PISTE, VFL_ICDE}, such as CGIR~\cite{CGIR} and \emph{PISTE}~\cite{PISTE}, have been proposed for HFL and VFL, respectively. These methods leverage a deep neural network (DNN) generator to assist in solving the regression task in Eq.~\eqref{eq:dra_obj}.
Recent work further demonstrate that a malicious server can actively manipulate the training process to steal client's data~\cite{loki, malious2, VFL_unleashing}.
In HFL, LOKI~\cite{loki} proposed to modify the model architecture and the parameters sent to clients to facilitate the attack. However, in practice, FL participants usually agree on the model architecture in advance, making abnormal model structures easy to detect and such attacks ineffective.
In the context of VFL, FSHA~\cite{VFL_unleashing} also introduced a DNN capable of mapping $\mathcal{F}(\bm{x})$ directly to $\hat{\bm{x}}$. Although this methodology facilitates effective reconstruction, it disrupts the conventional VFL training paradigm, thereby substantially degrading the performance of the classifier trained within the system.

\noindent \textbf{Privacy Metrics for FL.}
Xu et al.~\cite{MI-estimator} analyzed the privacy risks in HFL by estimating the mutual information-based channel capacity between training data and shared model parameters during each training round. 
However, this bound is computationally infeasible for high-dimensional, non-linear model architectures due to their inherent complexity. While~\cite{MI-estimator} introduces a tractable approximation of channel capacity using data-space mutual information, this approximation fails to account for the correlation between training data and the model parameters trained on it. As a result, it does not effectively capture the variability in privacy risks across different federated learning systems with diverse datasets and architectures.
Noorbakhsh et al.~\cite{MI-estimator2} proposed \textsf{Inf\textsuperscript{2}Guard} to mitigate inference attacks to balance privacy protection and system utility. It is designed to minimize the mutual information between original data and feature embeddings, while enhancing the correlation between feature embeddings and class labels. However, this approach does not serve to characterize and quantify the feasibility of DRA attacks across different datasets and models. It does not develop a quantified assessment of the DRA risk. Furthermore, \textsf{Inf\textsuperscript{2}Guard} can not be adopted in VFL as VFL clients do not have class labels during training. Though it can be used in HFL training. Attackers in HFL only need to access model parameters for DRA, not embeddings. 
Guo et al.~\cite{VFL_Fisherinfo} approached the quantification of DRA risk using the Fisher information of shared feature embeddings in VFL. However, dFIL is limited by its dependence on unbiased data reconstruction estimators for DRA attacks, whereas most practical DRA attacks rely on biased estimators. Additionally, dFIL does not facilitate comparisons of DRA risks across various model architectures. 
These limitations highlight the need for a unified and theoretically-guaranteed assessment framework to compare DRA risks across a wide range of data sources and model architectures.
\section{Invertibility Loss (\InvLoss)} \label{sec:method}
We begin by formally introducing the concept of Invertibility Loss \InvLoss, which defines the attack objective in DRA attacks. Next, we establish an upper bound for \InvLoss, laying the groundwork for a theoretical analysis of the feasibility of DRA attacks in FL systems. Our theoretical investigation of \InvLoss is approached from two perspectives. First, we leverage the proposed upper bound to analyze how existing defense mechanisms—such as injecting data-agnostic noise perturbations or performing parameter pruning—can mitigate the feasibility of DRA attacks. 
Building on these insights, we propose adapting noise perturbation based on the spectral properties of the Jacobian matrix of the model gradients or feature embeddings. This approach significantly improves task utility while maintaining robust privacy protection.
Second, we introduce a risk estimation framework for DRA attacks, termed \InvRE, which is grounded in the theoretical analysis of \InvLoss. For FL tasks involving diverse data sources and model architectures, \InvRE provides a bounded quantification of risk, enabling consistent assessment of vulnerability to DRA attacks across different FL scenarios.

\noindent\textbf{Threat model.} 
In this work, we focus on the privacy leakage caused by DRA attacks in HFL and VFL systems.
Specifically, we consider an honest but curious server as an attacker, who  aims to reconstruct a target client’s private input images by exploiting the parameters shared during the FL process. The attacker adheres to the FL protocol, i.e., it does not interfere with the training procedure or modify the model architecture to facilitate the attack, but solely leverages the exposed parameters (e.g., gradients or embeddings) from the victim client.
In HFL, such attacks occur during training, where the attacker residing at the central server observes model updates submitted by the victim client, consistent with prior work~\cite{DLG, IG, CGIR}. 
In contrast, privacy attacks against VFL occur at the inference time, as demonstrated in~\cite{Ressfl, VFL_ICDE}. The attacker at the server analyzes embedding vectors produced by the client’s local model for joint prediction to reversely estimate the input data.

\begin{definition}[Invertibility Loss]\label{def:InvL}
Given a target data instance $\bm{x}$, $\mathcal{F}(\bm{x})$ denotes the shared model updates or feature embedding vectors in HFL or VFL respectively. Such shared information can be considered as a functional transformation of $\bm{x}$. We define the inverse transformation function $\mathcal{A}_{\bm{x}}$ mapping $\mathcal{F}(\bm{x})$ to the data space. A DRA attack can be considered to estimate $\mathcal{A}_{\bm{x}}$ minimizing the Invertibility Loss (\InvLoss) over $\bm{x}$, which gives:
\begin{equation}\label{eq:InvL}
\InvLoss_{\bm{x}}  = \min_{\bm{\mathcal{A}}_{\bm{x}}} \,\|\mathcal{A}_{\bm{x}}(\mathcal{F}(\bm{x})) - \bm{x}\|^2,
\end{equation}
where $\|\cdot\|$ denotes the Euclidean distance between the reconstructed data profile and the ground truth.  
\end{definition}
The value of $\InvLoss_{\bm{x}}$ reflects the minimal possible data reconstruction error using the optimal inverse transformation $\mathcal{A}_{\bm{x}}$. It quantifies the discrepancy between the reconstructed data and the original data. The definition of $\InvLoss_{\bm{x}}$ aligns with the objective function of gradient- or embedding-matching-based DRA attack methods, as given in Eq.~\eqref{eq:dra_obj}. Therefore, Eq.~\eqref{eq:InvL} is used to evaluate the effectiveness of DRA attacks, where a lower $\InvLoss_{\bm{x}}$ indicates a more effective attack and, consequently, a higher risk of DRA attacks. 
The detailed explanation of the equivalence between Eq.~\eqref{eq:InvL} and Eq.~\eqref{eq:dra_obj} is provided in Appendix~\ref{sec:Consistency}.

\subsection{The upper bound of \InvLoss}\label{subsec:upper_bound_InvL}

\noindent \textbf{Rank-$k$ optimal DRA attacker.}  
To derive a tight upper bound of $\InvLoss_{\bm{x}}$ for a rank-$k$ optimal attacker, we first locally linearize the transformation function $\mathcal{F}(\bm{x})$ via Taylor expansion.
We let $\mathcal{F}(\bm{x}) = \bm{G}_{\bm{x}} \bm{x} + o(\delta^2)$, where $o(\delta^2)$ denotes higher-order infinitesimals. For simplicity, we denote $o(\delta^2)$ as $C$ in subsequent discussions.
To estimate the inverse transform $\bm{A}_{\bm{x}}$, DRA attackers may leverage various generative models and optimization techniques. 
Regardless of the specific learning techniques used for attack, minimizing $\InvLoss_{\bm{x}}$ involves solving a linear regression problem. By employing the Cauchy-Swartz inequality, the bound for $\InvLoss_{\bm{x}}$ can be expressed as:

\begin{equation}\label{eq:rankkattacker1}
\begin{split}
\InvLoss_{\bm{x}} & \leq \|\bm{A}_{\bm{x}}\bm{G}_{\bm{x}}{\bm{x}} - \bm{x}\|^2 + C\\
&\leq \|(\bm{A}_{\bm{x}}\bm{G}_{\bm{x}} - \bm{I})\bm{x}\|^2 + C.\\
\end{split}
\end{equation}

With Eq.~\eqref{eq:rankkattacker1}, minimizing $\InvLoss_{\bm{x}}$ to estimate the inverse transform $\mathcal{A}_{\bm{x}}$ as defined in Eq.~\eqref{eq:InvL} is equivalent to solving the following linear least squares regression problem:

\begin{equation}\label{eq:rankattacker2}
\bm{A}_{\bm{x}}^{*} =\underset{\bm{A}_{\bm{x}}}{\argmin}\,\|\bm{A}_{\bm{x}}\bm{G}_{\bm{x}}-\bm{I}\|_{\mathrm{F}}^2,
\end{equation}
where $\bm{I}$ is the identity matrix, and $\| \cdot \|_{\text{F}}$ denotes the Frobenius norm of a matrix. 
In the optimal case, the DRA attacker can reach the lowest reconstruction error when $\bm{A}_{\bm{x}}^{*}$ is the Moore-Penrose generalized inverse of $\bm{G}_{\bm{x}}$: 
\begin{equation}\label{eq:mpinverse}
\bm{A}_{\bm{x}}^{*} = \bm{V}\bm{\Sigma}^{\dagger}\bm{U}^{\top},
\end{equation}
where $\bm{V}$ and $\bm{U}$ are the left and right singular vectors of $\bm{G}_{\bm{x}}$. The matrix $\bm{\Sigma}^{\dagger}$ is a diagonal matrix. 
Each of its diagonal elements computes the reciprocal of each non-zero singular value of $\bm{G}_{\bm{x}}$, while leaving the zeros in place. 

In practice, a weak DRA attacker can not achieve the optimal solution to the inverse transform and can only approximate $\bm{A}_{\bm{x}}^{*}$ up to a rank $k$, with $k$ no larger than the number of non-zero singular values of $\bm{G}_{\bm{x}}$. We thus define a \textit{rank-k optimal DRA attacker} on a given instance $\bm{x}$ as the one that obtains the best rank-$k$ approximation $\tilde{\bm{A}}_{\bm{x}}^{k*}$ to $\bm{A}_{\bm{x}}^{*}$:
\begin{equation}\label{eq:rankattacker3}
\tilde{\bm{A}}_{\bm{x}}^{k*} = \bm{V}_{1:k}\tilde{\bm{\Sigma}}^{\dagger}\bm{U}^{\top}_{1:k},
\end{equation} 
where $\bm{V}_{1:k}$ and $\bm{U}^{\top}_{1:k}$ defines the first $k$ columns of the right and left singular vectors of $\bm{G}_{\bm{x}}$, corresponding to its $k$ largest non-zero singular values of. $\tilde{\bm{\Sigma}}^{\dagger}$ is the truncated version of $\bm{\Sigma}^{\dagger}$ containing the reciprocals of the $k$ largest non-zero singular values of $\bm{G}_{\bm{x}}$ in its diagonal, while the remaining diagonal elements are set to zeros.

Using Def.~\ref{def:InvL}, we establish an upper bound of the data reconstruction error of a rank-$k$ optimal DRA attacker over the input instance $\bm{x}$ and the model $\mathcal{F}$, as stated in Theorem~\ref{theorem:Bound_InvL}. 
\begin{theorem}[\InvLoss for rank-$k$ attacker]\label{theorem:Bound_InvL} 
Given a normalized example $\bm{x}\in{\mathbb{R}^{m}}$ as the target of DRA and its corresponding shared parameters $\mathcal{F}(\bm{x})\in{\mathbb{R}^{p}}$, the \InvLoss for a rank-$k$ optimal DRA attacker is bounded as:
\begin{equation}\label{eq:Bound_InvL}
\InvLoss_{\bm{x}} \leq \sum_{i=k+1}^{d} (\bm{V}_{i}^{\top}\bm{x})^2 + C,
\end{equation}
where $k$ represents the rank of the approximation defined in Eq.~\eqref{eq:rankattacker3} derived by the rank-$k$ optimal attacker, and $d= \min\{m, p\}$. $\bm{V}_{i}$ denotes the $i$-th column of $\bm{V}$, denoting the singular vector corresponding to the $i$-th largest singular value. $\bm{V}_{i}^{\top}\bm{x}$ hence denotes the projection of $\bm{x}$ onto the singular vector $\bm{V}_i$. 
\end{theorem}

\subsection{The upper bound of \InvLoss with existing defenses}\label{subsec:explain_existing_defense}

With Theorem~\ref{theorem:Bound_InvL}, we can systematically explain the effectiveness of the existing defense strategies against DRA via the lens of \InvLoss. These defense methods can be broadly classified into two categories: ``\emph{Noise Perturbation}'' \cite{DLG, PriVFL, DPSGD} and ``\emph{Information Compression} (IC)''  \cite{defense_dropout, defense_pruning}-based defenses.

Noise Perturbation-based defenses involve adding noise to the raw features to obscure sensitive information, thereby enhancing data privacy protection~\cite{DLG, PriVFL, DPSGD, SoteriaFL}.
Depending on where the noise is injected, the noise perturbation-based defense mechanisms can be further categorized: (i) \emph{Data-Level Noise Perturbation (DNP)}, where noise is directly applied to input data before processing in HFL and VFL; (ii) \emph{Gradient-Level Noise Perturbation (GNP)}, where noise is added to the gradients exchanged during training in HFL; and (iii) 
\emph{Embedding-Level Noise Perturbation (ENP)}, where noise is incorporated into embedding vectors generated by the client’s bottom model in VFL.
IC-based defense aims to reduce the information exchanged between the client and the server, thereby reducing the risk of sensitive information leakage. Two prominent approaches within this category are \emph{Prune}~\cite{defense_pruning} and \emph{Dropout}~\cite{defense_dropout}. Prune involves selectively removing less important shared features. 
Dropout, on the other hand, randomly sets a subset of shared features to zero during communication.
While the main purpose of them is to improve communication efficiency~\cite{acgd_compression, unified_compression, 3pc_compression, defense_compression2} or prevent overfitting~\cite{defense_dropout}, Prune and Dropout have been adapted to reduce the risk of data leakage by limiting the exposure of critical information~\cite{DLG, Ressfl}. 
In the following, we extend Theorem~\ref{theorem:Bound_InvL} to the theoretical implications of DNP and GNP/ENP for \InvLoss in Theorem~\ref{theorem:Bound_InvL_DNP} and Theorem~\ref{theorem:Bound_InvL_GENP}, respectively.

\begin{theorem}[\InvLoss under DNP]\label{theorem:Bound_InvL_DNP}
Given a normalized instance $\bm{x}\in{\mathbb{R}^{m}}$, subjected to DNP-based defense, and the shared information $\mathcal{F}(\bm{x})$ accessed by the attacker, the \InvLoss of a rank-$k$ optimal DRA attacker is bounded as:
\begin{equation}\label{eq:Bound_InvL_DNP}
\InvLoss_{\bm{x}} \leq \sum_{i=k+1}^{d} (\bm{V}_{i}^{\top}\bm{x})^2 + \sum_{i=1}^{k} \frac{(\bm{V}_{i}^{\top} \bm{\epsilon})^{2}}{m} + C,
\end{equation}
where $\bm{\epsilon}$ is the injected noise perturbation to $\bm{x}$; $k$ denotes the rank of the estimation defined in Eq.~\eqref{eq:rankattacker3}.
\end{theorem}

Compared to $\InvLoss_{\bm{x}}$ without any defense, as described in Theorem~\ref{theorem:Bound_InvL}, Theorem~\ref{theorem:Bound_InvL_DNP} indicates that 
the scale of noise injected into the input data is a critical factor in the defense effectiveness of DNP.
Specifically, increasing the magnitude of the injected noise enlarges the upper bound in Eq.~\eqref{eq:Bound_InvL_DNP}, which
enhances the protection strength of DNP. This observation is supported by experimental results presented in Fig.~\ref{fig:hfl_DNP} and Fig.~\ref{fig:vfl_DNP} for HFL and VFL respectively.

\begin{theorem}[\InvLoss under GNP/ENP]\label{theorem:Bound_InvL_GENP}
Given a normalized instance $\bm{x}\in{\mathbb{R}^{m}}$ targeted by a DRA and the shared information $\mathcal{F}(\bm{x})$, the \InvLoss of a rank-$k$ optimal DRA attacker with GNP or ENP-based defense applied is bounded as:
\begin{equation}\label{eq:Bound_InvL_GENP}
\InvLoss_{\bm{x}}  \leq \sum_{i=k+1}^{d} (\bm{V}_{i}^{\top}\bm{x})^2 + \sum_{i=1}^{k} \frac{(\bm{U}_{i}^{\top}\bm{\epsilon})^{2}}{\sigma _{i}^{2}p} + C,
\end{equation}
where $\bm{\epsilon}$ represents the noise perturbation injected into $\mathcal{F}(\bm{x})$; $k$ denotes the rank of the estimation as defined in Eq.~\eqref{eq:rankattacker3}. $\sigma_{i}$ is the singular values of $\bm{G}_{\bm{x}}$, and $p$ is the number of the dimensions in the injected noise. $\bm{U}_i$ denotes the left singular vector of the Jacobian matrix of $\mathcal{F}(\bm{x})$ corresponding to the $i$-th largest singular value.
\end{theorem}

Theorem~\ref{theorem:Bound_InvL_GENP} demonstrates that increasing the magnitude of the injected noise applied to the gradients or embeddings in FL communication significantly improves the defense capability of GNP/ENP. We further verify this theoretical finding in the experimental results illustrated by Fig.~\ref{fig:hfl_GNP} and Fig.~\ref{fig:vfl_ENP} for HFL and VFL respectively.

For IC-based defenses, the lower bound of \InvLoss is typically increased by reducing the information shared with the attacker during FL communication. For instance, $\InvLoss_{\bm{x}} \geq \sum_{i=d-q+1}^{d}(\bm{V}_{i}^{\top}\bm{x})^2$, where $q$ represents the number of the pruned/dropped dimensions in the shared $\mathcal{F}(\bm{x})$ under IC-based defense methods.
Dropping more dimensions results in a larger lower bound, thereby providing stronger defense against DRA attacks. 
Applying IC-based defenses (such as pruning and dropout) is equivalent to setting certain dimensions of the Jacobian matrix $\bm{G}_{\bm{x}}$ to 0, thereby lowering its rank. This rank reduction limits the number of singular values available for DRA attacks, thus decreasing the risk of privacy leakage. 
Consistent experimental results demonstrating the privacy-enhancing effects of IC-based methods are shown in Fig.~(\ref{fig:hfl_prune})--(\ref{fig:hfl_dropout}) and Fig.~(\ref{fig:vfl_prune})--(\ref{fig:vfl_dropout}) for HFL and VFL, respectively.

\subsection{Adaptive noise perturbation-based defense} \label{sec:adp_defense}

\begin{figure}[t]
\centering{\includegraphics[width = 0.98\linewidth] {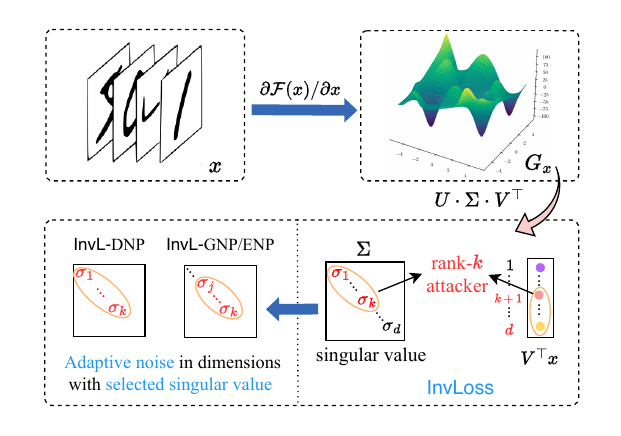}}
\caption{
Overview of the proposed \InvLoss and \InvL-based defenses.
\InvLoss measures reconstruction risk using key spectral components of the Jacobian $G_{\bm{x}}$, while \InvL-based defenses inject calibrated noise to these components to protect privacy while preserving model utility.}
\vspace{-3mm}
\label{fig:VFL-threat}
\end{figure}

Inspired by the upper bound analysis of \InvLoss provided in Theorem~\ref{theorem:Bound_InvL_DNP} and Theorems~\ref{theorem:Bound_InvL_GENP}, we propose two adaptive noise strategies that preserve protection strength while minimizing utility loss: \emph{Adaptive Data-level Noise Perturbation (\InvL-DNP)} and \emph{Adaptive Gradient/Embedding-level Noise Perturbation (\InvL-GNP or \InvL-ENP)}. Injecting the noise perturbation in the existing defense methods is to introduce perturbation blindly to the projection of the input data instance along all the singular vectors of the Jacobian matrix $\bm{G}_{\bm{x}}$. However, as demonstrated by Eq.~\eqref{eq:Bound_InvL_DNP} in Theorem~\ref{theorem:Bound_InvL_DNP} and Eq.~\eqref{eq:Bound_InvL_GENP} in Theorem~\ref{theorem:Bound_InvL_GENP}, the protection-enhancing effect of the injected noise $\bm{\epsilon}$ in DNP/GNP/ENP over the upper bound of \InvLoss is determined only by the projection of the noise $\bm{\epsilon}$ onto the singular vectors of $\bm{G}_{\bm{x}}$ corresponding to the largest $k$ singular values. On the contrary, the noise projected to the singular vectors with smaller singular values does not help increase the upper bound, thus contribute barely to improve the protection strength. 
Therefore, we propose to leverage this property to adaptively truncate the spectral projection of the injected noise perturbation. We strategically introduce noise only into the top-$k$ singular subspace to maintain the upper bound value of \InvLoss in Theorem~\ref{theorem:Bound_InvL_DNP}, hence maintaining the protection strength against DRA attacks and reducing the magnitude of the added noise.

The implementation steps for \InvL-DNP and \InvL-GNP/\InvL-ENP are organized as follows:

\textbf{Step 1.} Perform SVD of the Jacobian matrix $\bm{G}_{\bm{x}}$, $\bm{G}_{\bm{x}}= \bm{U} \bm{\Sigma} \bm{V}^{\top}$.

\textbf{Step 2.} Sample a random noise matrix $\bm{\epsilon}$ from a centered Gaussian distribution $\mathcal{N}(0, \delta)$. $\delta$ denotes the predefined variance.

\textbf{Step 3.} Project $\bm{\epsilon}$ onto the singular vectors of $\bm{G}_{\bm{x}}$, yielding $\bm{n} = \bm{V}^{\top}\bm{\epsilon}$ for \InvL-DNP, and $\bm{n} = \bm{U}^{\top}\bm{\epsilon}$ for \InvL-GNP/\InvL-ENP. Furthermore, we set the noise projection corresponding to the all singular values $\sigma_{i}$ except from the largest $k$ ones as zeros, i.e., $\bm{n}[k:] =0$. 
In our experimental study, we empirically find that setting $k$ to above the top 95\% of the sum of all singular values can strike a balance between the defense performance and its impact on the main task of FL.

Moreover, for \InvL-GNP/\InvL-ENP, according to Eq.~\eqref{eq:Bound_InvL_GENP} in Theorem~\ref{theorem:Bound_InvL_GENP}, the noise projection onto the singular vectors corresponding to the leading $k$ singular values has a diminishing magnitude with increasingly larger singular values $\sigma_i$. Therefore, we can additionally set the noise projection dimensions corresponding to the top $j$ ($j<k$) singular vectors to zero to reduce the utility loss. 
In our experiments, we empirically set $j$ to the top 60\% of all singular values, ensuring a balance between reducing the impact of noise and maintaining practicality.

\textbf{Step 4.} Map the truncated noise projection $\bm{n}$ back to the original data space to derive the adaptive noise, i.e., $\hat{\bm{\epsilon}} = \bm{V}\bm{n}$ for \InvL-DNP, and $\hat{\bm{\epsilon}} = \bm{U}\bm{n}$ for \InvL-GNP/\InvL-ENP.

We add the adaptive noise $\hat{\bm{\epsilon}}$ to the original data matrix $\bm{x}$ to produce the perturbed data, $\hat{\bm{x}} = \bm{x} + \hat{\bm{\epsilon}}$. For \InvL-GNP, we introduce the adaptive noise $\hat{\bm{\epsilon}}$ to the model gradient in each round of HFL training. For \InvL-ENP, we add the adaptive noise to the feature embedding produced by the bottom model of the target client in VFL. The noise-perturbed feature embedding is then fed to the top-level model for inference. Beyond the theoretical reasoning, we provide the experimental results of \InvL-DNP and \InvL-GNP/\InvL-ENP in Table~\ref{table:HFL_defense} and Table~\ref{table:VFL_defense} for HFL and VFL systems, respectively. The results confirm that FL systems deployed with the adaptive noise perturbation methods reach higher classification accuracy yet maintain the same or even stronger protection strength, compared to the vanilla DNP and GNP/ENP-based defense. 
To facilitate scalable computation, for \InvL-GNP, we perform the SVD decomposition on the averaged Jacobian matrix with respect to the class centers to derive the adaptive noise.

\subsection{Estimating the risk of DRA}
Based on the upper bound of \InvLoss in Theorem~\ref{theorem:Bound_InvL}, we propose an estimator to the risk of DRA attacks for a given FL system. We consider evaluating the risk under varying levels of attack strength, in order to make the risk estimation independent from the choice of DRA attack methods. We hereafter name the proposed quantified risk evaluator as \InvRE (Invertibility-based Risk Estimator).

According to Theorem~\ref{theorem:Bound_InvL}, increasingly larger $k$ indicates a stronger rank-$k$ optimal attacker. On one hand, with a higher rank $k$, more singular vectors are used by the attacker to compute the Moore-Penrose inverse to the Jacobian matrix, thus producing more accurate data reconstruction as suggested in Eq.~\eqref{eq:Bound_InvL}. On the other hand, uncovering more singular vectors increases the computational challenge to the attack method from two perspectives. \textbf{First}, singular vectors with marginally small singular values are difficult to derive due to numerical instability. Small singular values are close to zero, meaning their reciprocals used in the pseudo-inverse calculation become excessively large, which amplifies any small errors in the SVD results. This leads to a loss of numerical precision, as even tiny perturbations in the data or rounding errors can disproportionately affect the computed results. As a result, the singular vectors associated with these small singular values become unreliable, making their accurate recovery computationally difficult and sensitive to perturbations. \textbf{Second}, a small gap between successive singular values makes it difficult to distinguish the corresponding singular vectors. In such cases, these singular vectors can overlap or become nearly indistinguishable. Small changes to the Jacobian matrix, such as rounding errors or noise, can hence cause significant shifts in the orientations of such singular vectors. \textbf{In summary}, smaller magnitudes of the singular values and smaller singular value gap reduce the computational feasibility of exploiting all $k$ singular vectors for data reconstruction, thus making the corresponding rank $k$ attack less likely to conduct. 

Integrating all these considerations, we formulate \InvRE in Def.~\ref{def:E_InvL} using the averaged upper bound of \InvLoss, weighed by the singular value magnitudes and the gap between singular values of the Jacobian matrix.

\begin{definition}[\InvRE estimator of DRA risk]~\label{def:E_InvL}
Given a normalized data instance $\bm{x} \in \mathbb{R}^{m}$ 
and the corresponding shared parameters $\mathcal{F}(\bm{x}) \in \mathbb{R}^{p}$, the risk estimator for DRA attacks over $\bm{x}$ is given as:
\begin{equation}\label{eq:E_InvL}
\begin{split}
& \InvRE_{\bm{x}} = \frac{1}{1+\exp{(\beta(\sum_{k=1}^{d} P_k\tau_{k}}-\alpha))}, \\
& \text{where} \ P_{k} =\frac{1/T_{k}}{\sum_{j=1}^{d}(1/T_{j})} \ \text{with} \ T_{k}= \sum_{i=1}^{k} \frac{\sigma_i}{\sigma_i - \sigma_{i+1}}\\
& \text{and} \ \tau_{k} = \sum_{k+1}^{d}(\bm{V}^{\top}\bm{x})^2,
\end{split}
\end{equation}
where $\bm{G}_{\bm{x}}$ denotes the Jacobian matrix. 
$d=\min\{m,p\}$, $P_{k}$ quantifies the likelihood of successfully reconstructing the singular vectors of the largest $k$ singular values of $\bm{G}_{\bm{x}}$. $\tau_k$ represents the upper bound of \InvLoss with respect to a rank-k optimal attacker and given $\bm{x}$ and $\mathcal{F}$. $\alpha$ is the empirical mean of the values of $\sum_{k=1}^{d}P_{k}\tau_{k}$ derived over various datasets. $\beta$ is a scaling factor chosen empirically to avoid saturated values in the softmax function. In practice, we set $\beta$ to 5.
\end{definition}

\noindent \textbf{Properties of \InvRE.} 
\textbf{First}, \InvRE is computed in an agnostic way to specific attack methods $\mathcal{A}_{\bm{x}}$. 
Instead of choosing a concrete attack strategy for the risk assessment,
Eq.~\eqref{eq:E_InvL} takes a weighted average of $\tau_k$ to assess the overall DRA risk with respect to different attack strength, varied from weak to strong DRA attackers.  
$P_{k}$ takes the quotient of each singular value $\sigma_i$ ($i\leq{k}$) and the gap between successive singular values $\sigma_i$ and $\sigma_{i-1}$. As discussed before, a higher/lower $\sigma_i$ with a smaller/larger gap  indicates more/less difficult to reach the rank-$k$ optimal DRA attacks respectively. Therefore, we assign a higher/lower $P_k$ accordingly to the estimated invertibility loss \InvLoss of the corresponding rank-$k$ optimal attacker. 
With this setting, \InvRE produces a DRA risk estimation weighed by the computational feasibility of DRA attacks of different strength levels. More feasible DRA attacks with higher $P_k$ are more prevalent in practices, as such attack scenarios are easier to deliver. Hence these attack scenarios are given higher weights in the computation of \InvRE. On the contrary, stronger attacks with higher $k$ are in general less feasible. Though they produce better data reconstruction accuracy, they are assigned with lower weights in \InvRE. 

\textbf{Second}, Eq.~\eqref{eq:E_InvL} provides an instance-specific estimation of DRA risk. This allows users of FL systems to understand how data leakage risks vary across different data instances $\bm{x}$. Downstream FL applications can apply stricter protection measures to data instances with higher information leakage risks while relaxing controls for those with lower risks, thereby minimizing unnecessary perturbations that might affect classification performance. Empirically, we demonstrate this instance-wise DRA risk variation in both HFL and VFL systems using LeNet and ResNet architectures, as shown in Fig.~\ref{fig:cross_samples}. We reveal that the variations in \InvRE-based risk estimation align closely with the practical attack performance across different instances, validating \InvRE's effectiveness in evaluating instance-specific DRA risks.

\textbf{Third}, \InvRE does not pose any assumptions about the model architecture used in the target FL systems. \InvRE can thus be applied to arbitrary DNN models and measure the variation of the DRA risk determined by different model architecture. This facilitate the FL users to choose appropriate models in the target FL systems to mitigate the data leak risk. 
The experimental results, as shown in Fig.~\ref{fig:HFL-cross-models-mse} and Fig.~\ref{fig:VFL-cross-models-mse} for HFL and VFL systems, confirm that \InvRE-based risk estimation aligns consistently with the reported DRA attack performance metrics across DNN model architectures of varying complexities. These findings validate \InvRE's capability of accurately capturing variations of the DRA risk across different model structures used in FL systems.

\section{Experiments} \label{sec:experiment}
In this section, we present a comprehensive empirical evaluation to validate the \InvRE-based DRA risk estimator across 4 datasets from diversified applications and 3 model architectures with varying complexities used in HFL and VFL systems. Furthermore, we assess the defense effectiveness of the proposed adaptive noise perturbation-based protection scheme against DRA. Our empirical study is dedicated to providing systematic answers to the following questions:   

\textbf{\emph{Q1:}} Can \InvRE quantify the DRA risk associated with different data instances in both HFL and VFL frameworks? 

\textbf{\emph{Q2:}} Can \InvRE measure the DRA risk associated with different model structures adopted in FL systems with different DNN architectures? 

\textbf{\emph{Q3:}} 
Can the theoretical upper bound of \InvLoss in Sec.~\ref{subsec:explain_existing_defense} explain the effectiveness of existing different defense mechanisms against DRA in FL systems?  

\textbf{\emph{Q4:}} How can the \InvRE risk estimator be practically applied to guide client-side decision making in FL systems?

\textbf{\emph{Q5:}} Can the proposed adaptive noise perturbation mechanism improve the trade-off between model utility and data privacy protection compared to state-of-the-art noise perturbation and information compression defense methods in HFL and VFL?

The answer to \textbf{\emph{Q1}} and \textbf{\emph{Q2}} can provide empirical observations to verify whether the proposed \InvRE risk estimator can reflect the effectiveness of DRA attacks, thereby capturing the variation in DRA risk levels across different data instances and model architectures. 
Empirical observations related to \textbf{\emph{Q3}} further confirm that \InvRE can effectively assess DRA risk under different deployed defense mechanisms.
Based on the above three questions, \textbf{\emph{Q4}} investigates the practical implications of the \InvRE risk estimator in guiding client-side decision making within FL systems.
\textbf{\emph{Q5}} evaluates the overall effectiveness of the proposed adaptive noise perturbation mechanism, particularly its ability to maintain high utility of the FL models while offering strong protection against data reconstruction attacks.
All experiments are conducted on 2 servers: one with a 20-core Intel Xeon Platinum 8457C CPU and a NVIDIA T20 GPU (48 GB), and another with 160-core Intel Xeon Gold 6230 CPUs and 4 NVIDIA RTX 4090 GPUs (24 GB each).

\subsection{Experimental settings} \label{subsec:Experiment_Settings}
\noindent \textbf{Datasets.} 
Our experiments are trained on 4 real-life classical datasets: SVHN~\cite{SVHN}, CIFAR10~\cite{CIFAR10}, STL10~\cite{stl10}, and Tiny-ImageNet with 4-class classification~\cite{ImageNet} (abbreviated as ImageNet). These datasets have been widely used in the study of DRAs against FL systems~\cite{MI-estimator, CGIR, VFL_unleashing}. All dataset samples are resized to 32×32 pixels, a standard preprocessing step in ML to ensure consistency and fairness when comparing across different datasets~\cite{ML-docotor}.

\noindent \textbf{Model architectures of FL.} 
For HFL, we follow the experimental setup described in~\cite{DLG} and conduct experiments using three classical model architectures for image-domain tasks: LeNet~\cite{lenet}, AlexNet~\cite{alexnet}, and Residual network~\cite{resnet} (ResNet).
For VFL, we follow the configurations outlined in~\cite{VFL_Fisherinfo, VFL_unleashing, Ressfl} and examine three cut-layer configurations of ResNet18 to generate varying depth levels for the local client.
These configurations include cutting early (after the first block), at an intermediate depth (after the second block), and late (after the third block), referred to as ResNet-cut1, ResNet-cut2, and ResNet-cut3, respectively.
For all experiments, we set the mini-batch size to 32 and use cross-entropy as the loss function. The optimization process is carried out using the Adam optimizer with a learning rate of $1 \times 10^{-4}$. Each target model is trained within 200 epochs to ensure the convergence of the classifier.

\noindent \textbf{DRA methods and evaluation metrics.} 
To assess whether \InvRE can accurately indicate information leakage risks, we implement three representative attack methods conducted by an honest but curious server: 1).~\emph{DLG}, designed for HFL, minimizes the L2 distance between the gradient (HFL) or embedding (VFL) of the true input data and the reconstructed data; 2).~\emph{IG}, which improves upon \emph{DLG} by using cosine distance for gradient or embedding matching and incorporating total variation (TV) as a prior to reduce artifacts in the reconstructed data; and 3) \emph{CGIR} or \emph{PISTE} for HFL and VFL respectively, which enhance reconstruction capabilities using a deep generative model. 
The implementations of these attacks were based on the official code repositories provided by the respective authors~\cite{DLG, IG, CGIR, PISTE}. For \emph{DLG}, the L-BFGS optimizer was employed with 500 optimization iterations across all datasets. For \emph{IG}, the Adam optimizer was used with 24,000 optimization iterations, while for \emph{CGIR} and \emph{PISTE}, the Adam optimizer was applied with 2,000 optimization iterations. 
For demonstrating the attack with a malicious server, we involve LOKI~\cite{loki} and provide the empirical analysis of LOKI-driven attacks in Appendix~\ref{sec:more_results_loki}.

To quantify the relationship between the \InvRE score and the corresponding attack performance, we report both the Pearson correlation coefficient and the p-value (denoted as $\textsf{p}$) to assess the statistical significance of the alignment between the \InvRE score and the attack effectiveness. A p-value below 0.05 is widely considered indicative of a statistically significant correlation between the variables~\cite{Spearman}. Therefore, a higher Pearson correlation coefficient would validate \InvRE's ability to reflect DRA risks accurately. We adopt the standard classification performance metric, i.e., \emph{Accuracy} ({ACC}), to evaluate the utility of FL systems~\cite{ML-docotor}.
We evaluate the attack performance of involved DRA methods using widely adopted reconstruction quality metrics, including of Mean Square Error ({MSE}), Peak Signal-to-Noise Ratio ({PSNR}), and Structural Similarity Index Metric ({SSIM}) for the reconstructed data~\cite{IG}.

\noindent \textbf{Defense mechanisms.} 
We evaluate the performance of \InvRE score and 4 state-of-the-art defense methods tailored to HFL and VFL.
For HFL, we include \emph{Prune}\cite{defense_pruning}, \emph{Dropout}\cite{defense_dropout}, \emph{DNP}\cite{Ressfl}, and \emph{GNP}\cite{DLG}; for VFL, the evaluated methods are \emph{Prune}, \emph{Dropout}, \emph{DNP}, and \emph{ENP}\cite{DLG}.
\emph{Prune} sets the smallest $\lambda\%$ of elements to zero while retaining the rest\cite{Ressfl}, whereas \emph{Dropout} randomly zeros out $\lambda\%$ of elements.
\emph{DNP} introduces Gaussian noise in the original data space, while \emph{GNP} (for HFL) and \emph{ENP} (for VFL) apply similar noise to gradients or feature embeddings, respectively.
Note that, we do not consider encryption-based defense strategies~\cite{defense_HE, defense_SS}, such as Falcon~\cite{defense_falcon} and HashVFL~\cite{Hashvfl}, due to their significant communication and computation overhead.

\subsection{Evaluating \InvRE for different instances} 
In this section, we aim to answer $\textbf{\emph{Q1}}$. We compute both the Pearson correlation coefficient $\textsf{cor}$ and the corresponding p-value between the \InvRE score and the corresponding reconstruction MSE derived by different attacks over each input instance. A detailed description is provided in Appendix~\ref{sec:E_mse}.

\begin{figure}[t]
    \centering
    \begin{subfigure}[b]{0.48\textwidth}
        \centering
        \includegraphics[width=\textwidth]{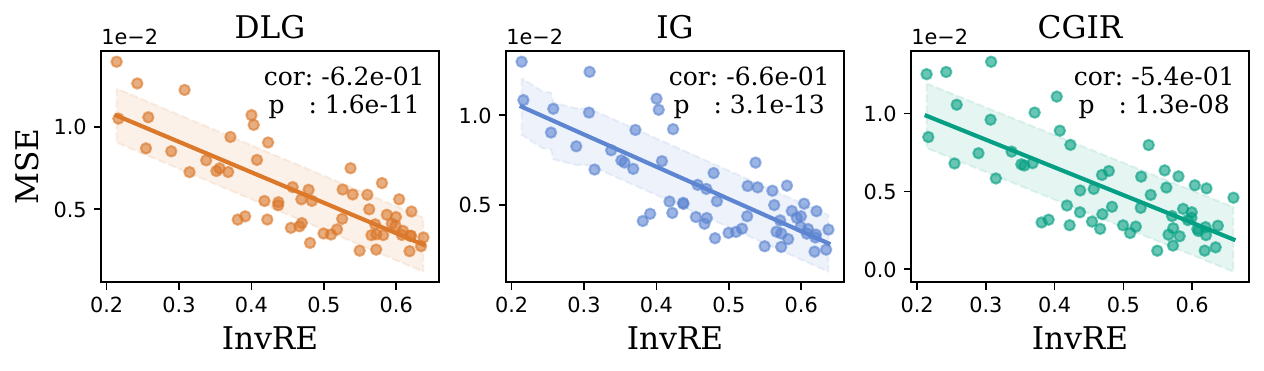} 
        \caption{HFL}
        \label{fig:A_HFL_CGIR-cifar10_lenet_privacy_mse_scatter}
    \end{subfigure}
    \begin{subfigure}[b]{0.48\textwidth}
        \centering
        \includegraphics[width=\textwidth]{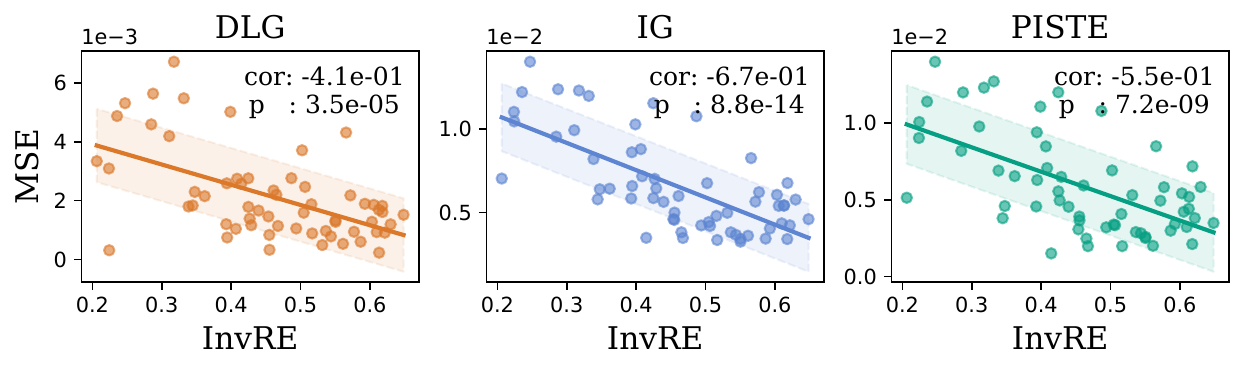} %
        \caption{VFL}
        \label{fig:A_VFL_CGIR-cifar10_resnet1_privacy_mse_scatter}
    \end{subfigure}
    \vspace{-5mm}
    \caption{Correlation between \InvRE and reconstruction MSE for different samples in CIFAR10 trained on the LeNet and ResNet-cut1 for HFL and VFL respectively.}
    \label{fig:cross_samples}
\end{figure}

\begin{figure}[t]
\centering{\includegraphics[width = 0.7\linewidth] {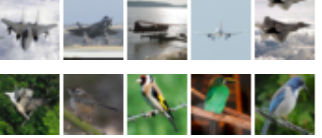}}
\caption{The five images with the smallest (top row) and largest (bottom row) \InvRE values over STL10.}
\label{fig:HFL_visual_topk}
\end{figure}

Fig.~\ref{fig:cross_samples} provide the \InvRE score and corresponding reconstruction MSE under three distinct attack methods for CIFAR10 trained on the LeNet and ResNet-cut1
of HFL and VFL. 
Due to space limitations, comprehensive results on other three datasets are provided in Appendix~\ref{sec:more_results} (Fig.~\ref{fig:HFL_cross_samples} and Fig.~\ref{fig:VFL_cross_samples}). 
To further demonstrate the impact of intrinsic image characteristics on \InvRE scores, we show the five original images with the lowest and highest scores for HFL and VFL in Fig.~\ref{fig:HFL_visual_topk} and Fig.~\ref{fig:VFL_visual_topk} (Appendix~\ref{sec:more_results}), respectively.

Fig.~\ref{fig:cross_samples} demonstrates a significant negative correlation between \InvRE score and data reconstruction MSE across various datasets (CIFAR10, STL10, SVHN, and ImageNet) for both HFL and VFL frameworks, with correlation coefficients ranging from -0.41 to -0.67 and p-values below $1e^{-5}$. The consistent negative correlation observed across different DRA attack methods (DLG, IG, and CGIR/PISTE) highlight \InvRE's adaptability in reflecting the data privacy risk level across diverse FL applications.
In addition, as shown in Fig.~\ref{fig:HFL_visual_topk}, images with the lowest \InvRE scores (top rows) tend to have simpler and smoother textures, leading to lower gradient and feature sensitivity to input variations. This makes it harder for attackers to reconstruct the original data, indicating higher robustness to DRAs. In contrast, images with the highest \InvRE scores (bottom rows) often have more complex and distinctive textures, making their gradients and embeddings more sensitive and thus easier to exploit for reconstruction. 
This highlights the key role of feature representation in determining and mitigating the DRA risk over different data instances, as discussed in \cite{MI-estimator2}. 

\textbf{Summary}: The experimental results underscore that \InvRE serves as a reliable metric for quantifying privacy leakage risks associated with different samples in both HFL and VFL frameworks. The consistency of the observed correlation patterns across different datasets and attack methods enhances its practicality in real-world FL systems.

\begin{figure*}[t]
\centering{\includegraphics[width = 0.9\linewidth] {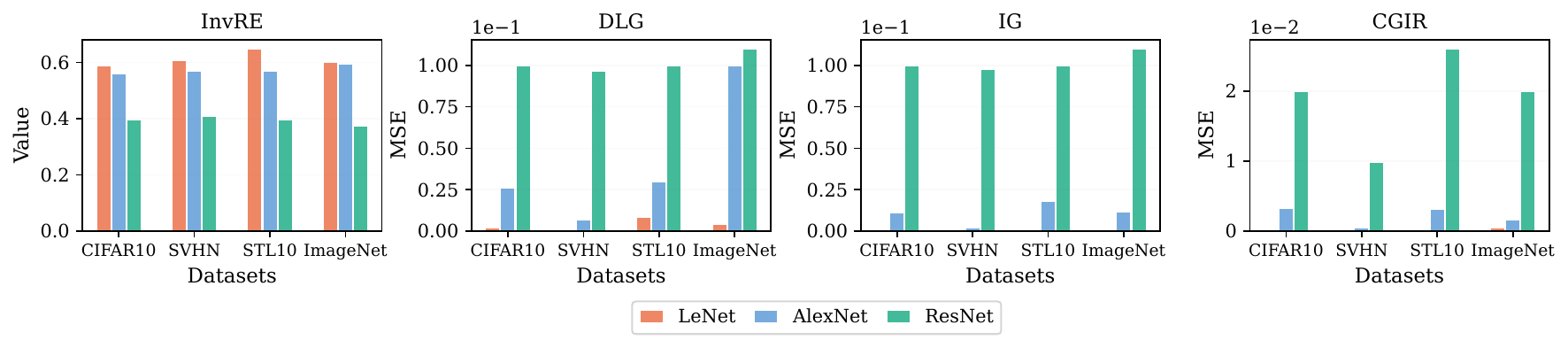}}
\vspace{-2mm}
\caption{Correlation between \InvRE and reconstruction error (MSE) for different model structures in HFL.}
\label{fig:HFL-cross-models-mse}
\end{figure*}
\begin{figure*}[h]
\centering{\includegraphics[width = 0.9\linewidth] {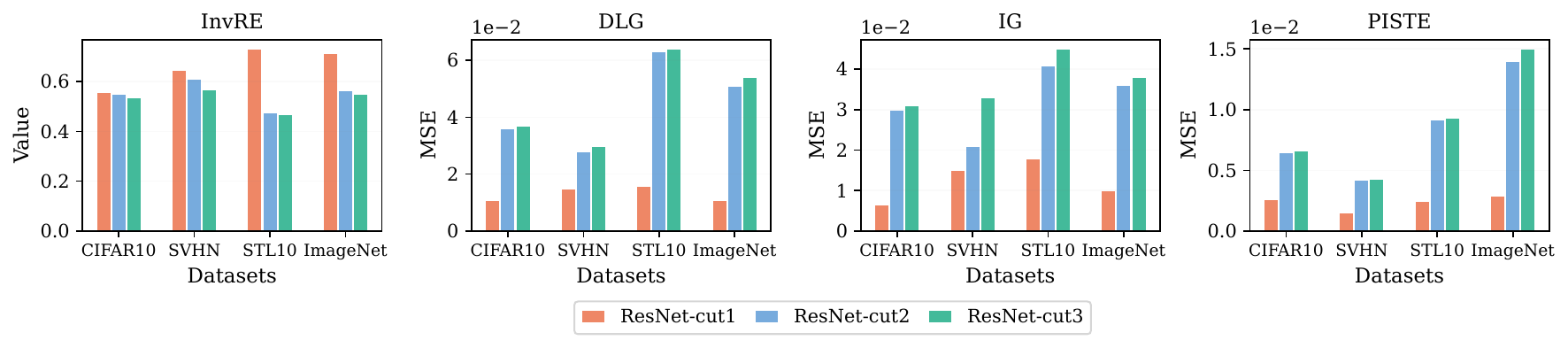}}
\vspace{-2mm}
\caption{Correlation between \InvRE and reconstruction error (MSE) for different model structures in VFL.}
\label{fig:VFL-cross-models-mse}
\end{figure*}

\subsection{Evaluating \InvRE cross different model structures in FL}\label{subsec:explain_models}

This section addresses $\textbf{\emph{Q2}}$: We evaluate the correlation between \InvRE score and the reconstruction MSE using various model architectures deployed in HFL and VFL systems across four datasets: CIFAR10, SVHN, STL10, and ImageNet.
The results are presented in Fig.~\ref{fig:HFL-cross-models-mse} and Fig.~\ref{fig:VFL-cross-models-mse} for HFL and VFL, respectively.

The experimental results demonstrate a significant correlation between \InvRE score and the reconstruction errors across different model structures in both HFL and VFL. 
Within the HFL framework, the LeNet model, characterized by a higher \InvRE score, exhibits greater vulnerability to DRA attack, as indicated by its lower MSE values. For instance, on the STL10 dataset under the DLG attack, LeNet achieves an MSE of 0.008, which is significantly lower than the values for AlexNet (0.026) and ResNet (0.1). Similarly, in the VFL framework, ResNet-cut1, having a higher \InvRE score, encounters a greater risk of DRA compared to ResNet-cut2 and ResNet-cut3.
This trend can be explained by the higher complexity and non-linearity of deeper models, such as AlexNet and ResNet, which make it more difficult for DRA attacks to invert transformations and reconstruct data.

\textbf{Summary.} The experimental findings confirm that \InvRE score serves as a robust indicator of privacy risks across different model architectures in both HFL and VFL systems. With \InvRE, FL practitioners can evaluate privacy risks and optimize model configurations before FL training, enabling more secure and privacy-preserving FL deployment. 

\begin{figure}[t!]
    \centering
    \begin{subfigure}[b]{0.46\textwidth}
        \centering
        \includegraphics[width=\textwidth]{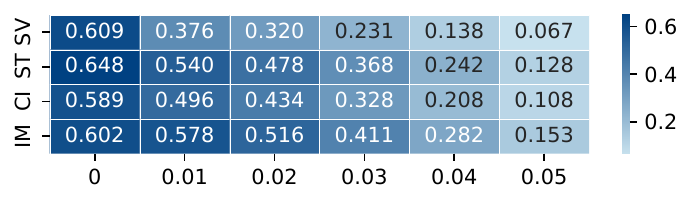} %
        \caption{DNP}
        \label{fig:hfl_DNP}
    \end{subfigure}
    \begin{subfigure}[b]{0.46\textwidth}
        \centering
        \includegraphics[width=\textwidth]{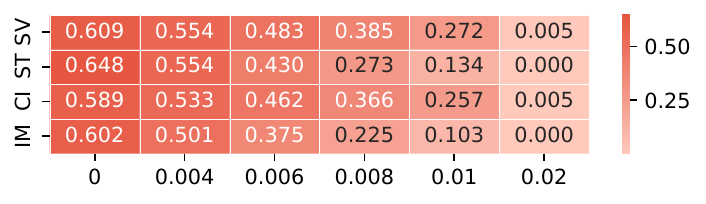} %
        \caption{GNP}
        \label{fig:hfl_GNP}
    \end{subfigure}
    \begin{subfigure}[b]{0.46\textwidth}
        \centering
        \includegraphics[width=\textwidth]{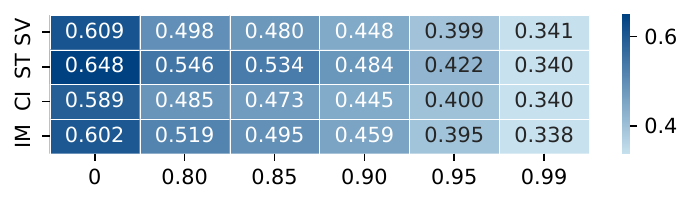} 
        \caption{Prune}
        \label{fig:hfl_prune}
    \end{subfigure}
    \begin{subfigure}[b]{0.46\textwidth}
        \centering
        \includegraphics[width=\textwidth]{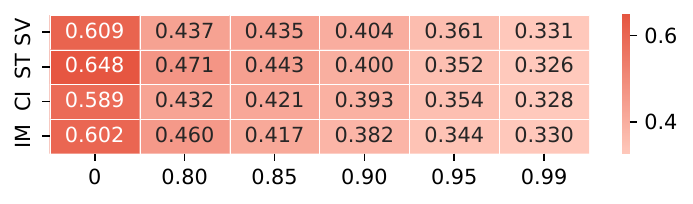} %
        \caption{Dropout}
        \label{fig:hfl_dropout}
    \end{subfigure}
    \caption{Values of \InvRE with LeNet under different defense strategies and strengths in HFL.}
    \label{fig:explain-defense-HFL}
\end{figure}


\begin{figure}[t!]
    \centering
    \vfill
    \begin{subfigure}[b]{0.46\textwidth}
        \centering
        \includegraphics[width=\textwidth]{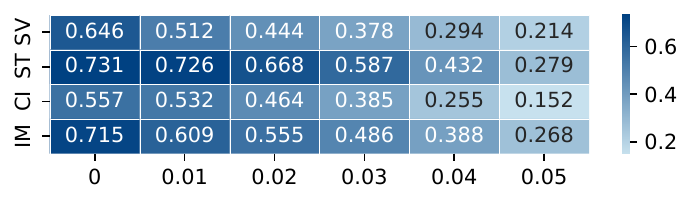} %
        \caption{DNP}
        \label{fig:vfl_DNP}
    \end{subfigure}
    \begin{subfigure}[b]{0.46\textwidth}
        \centering
        \includegraphics[width=\textwidth]{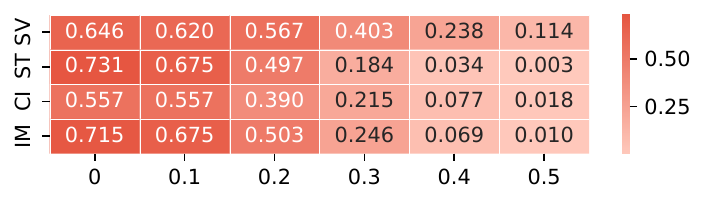} %
        \caption{ENP}
        \label{fig:vfl_ENP}
    \end{subfigure}    
    \begin{subfigure}[b]{0.46\textwidth}
        \centering
        \includegraphics[width=\textwidth]{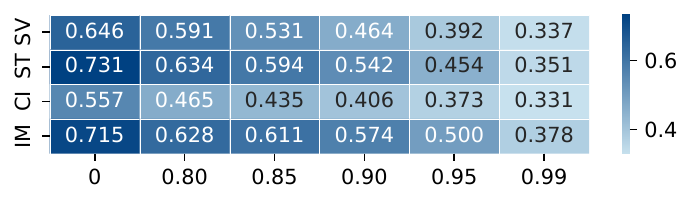} 
        \caption{Prune}
        \label{fig:vfl_prune}
    \end{subfigure}
    \begin{subfigure}[b]{0.46\textwidth}
        \centering
        \includegraphics[width=\textwidth]{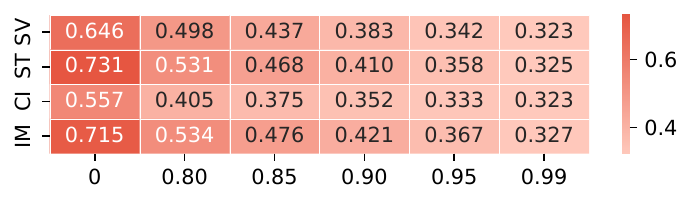} %
        \caption{Dropout}
        \label{fig:vfl_dropout}
    \end{subfigure}
    \caption{Values of \InvRE with ResNet-cut1 under different defense strategies and strengths in VFL.}
    \label{fig:explain-defense-VFL}
\end{figure}

\subsection{Evaluating defense methods with \InvRE}
Based on the answer to \textbf{\emph{Q1}} and \textbf{\emph{Q2}}, we further utilize \InvRE score to demonstrate the reduction of DRA risk after deploying the state-of-the-art defense mechanisms, including the noise perturbation and information compress-based methods. Through empirical results, we show that the theoretical analysis over the upper bound of the \InvLoss forms the foundation for explaining the effectiveness of these defense methods. It provides the answer to \textbf{\emph{Q3}}.
Fig.~\ref{fig:explain-defense-HFL} and Fig.~\ref{fig:explain-defense-VFL} present the \InvRE score with DNP and GNP/ENP, as well as Prune and Dropout for LeNet in HFL and ResNet-cut1 in VFL respectively. 
Additional models are provided in Appendix~\ref{sec:more_results}.
In these figures, SV, ST, CI, and IM denote the SVHN, STL10, CIFAR10, and ImageNet datasets, respectively. The horizontal axis indicates the increase in defense strength as the defense parameter grows.
The first column in each sub-figure shows the \InvRE score without the defense methods deployed, labeled as `0'.

By analyzing the results presented in Fig.~\ref{fig:explain-defense-HFL} and Fig.~\ref{fig:explain-defense-VFL}, it is evident that increasing the defense strength, whether through noise perturbation methods (DNP and GNP/ENP) or information compression methods (Prune and Dropout), significantly reduces privacy risks, as measured by the \InvRE score. 
Specifically, noise perturbation-based defenses (DNP and GNP/ENP) achieve this reduction by introducing noise to the raw features, thereby disrupting the attacker's ability to reconstruct data features. For example, as shown in Fig.~\ref{fig:explain-defense-HFL}, deploying the GNP method in the HFL setting with LeNet results in a significant reduction in the \InvRE score as the noise variance increases from 0.004 to 0.02. 
On the other hand, information compression-based defenses (Prune and Dropout) mitigate privacy leakage by discarding a portion of the shared information. As shown in Fig.~\ref{fig:explain-defense-VFL}, the Prune defense method applied to ResNet-cut1 in the VFL setting achieves a steady reduction in the \InvRE score as the discarding ratio increases from 0.80 to 0.99.
Therefore, it is essential to design more advanced defense strategies that can better balance privacy protection and model utility.

\textbf{Summary.} From the empirical results, we can observe consistently the descending tendency of \InvRE score with stronger defense strength using the state-of-the-art defense mechanisms and over different datasets. With the empirical observation, we can find that the proposed theoretical upper bounds of the \InvLoss in Eq.~\eqref{eq:Bound_InvL_DNP} and Eq.~\eqref{eq:Bound_InvL_GENP} provide a consistent explanation to the effectiveness of the state-of-the-art defensive mechanisms. 
The results show that the theoretically guaranteed \InvRE score can be used to systematically evaluate the effectiveness of defense mechanisms.

\begin{figure}[t!]
    \centering
    \begin{subfigure}[b]{0.35\textwidth}
        \centering
        \includegraphics[width=\textwidth]{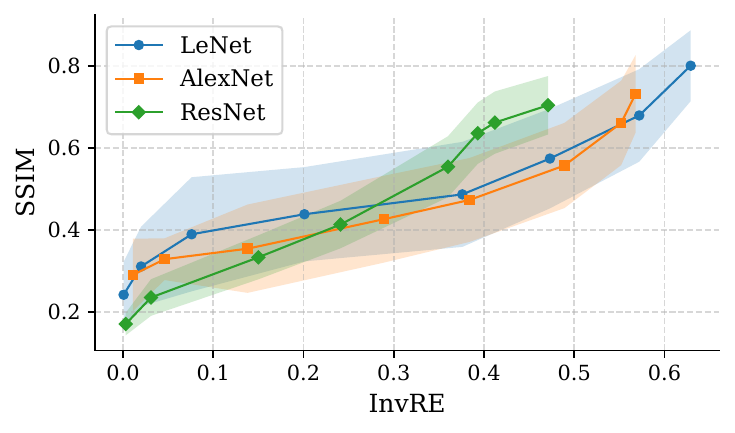} %
        \label{fig:Invre_hfl_SSIM}
    \end{subfigure}
    \begin{subfigure}[b]{0.4\textwidth}
        \centering
        \includegraphics[width=\textwidth]{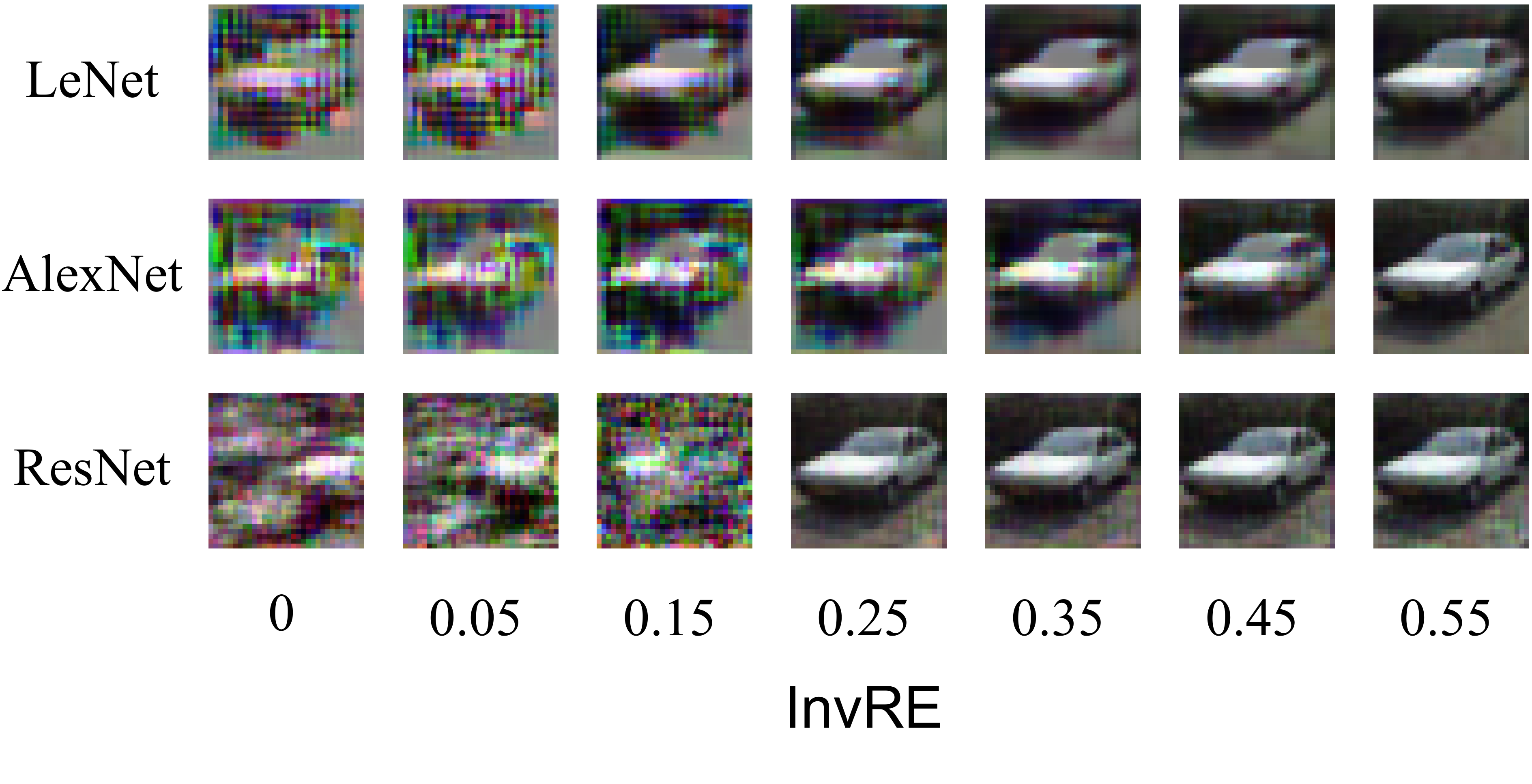} %
        \label{fig:Invre_hfl_imgs}
    \end{subfigure} 
    \vspace{-7mm}
    \caption{\InvRE vs. reconstruction quality in HFL. Top: reconstructed images across increasing \InvRE levels; bottom: corresponding SSIM scores.}
    \label{fig:Invre_hfl}
\end{figure}

\begin{figure}[h]
    \centering
    \begin{subfigure}[b]{0.35\textwidth}
        \centering
        \includegraphics[width=\textwidth]{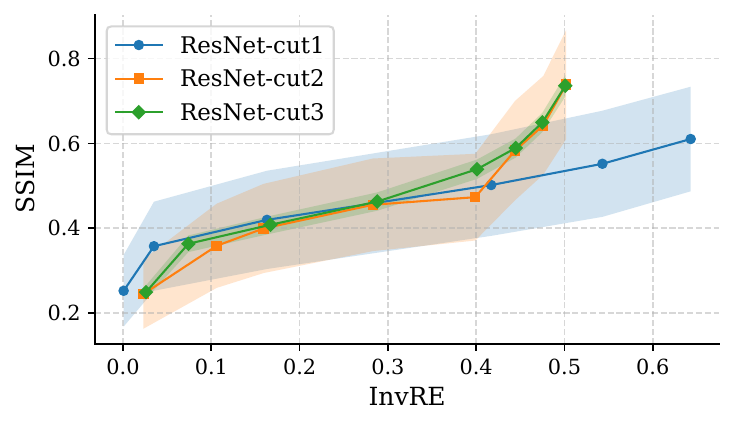} %
        \label{fig:Invre_vfl_SSIM}
    \end{subfigure}
    \begin{subfigure}[b]{0.4\textwidth}
        \centering
        \includegraphics[width=\textwidth]{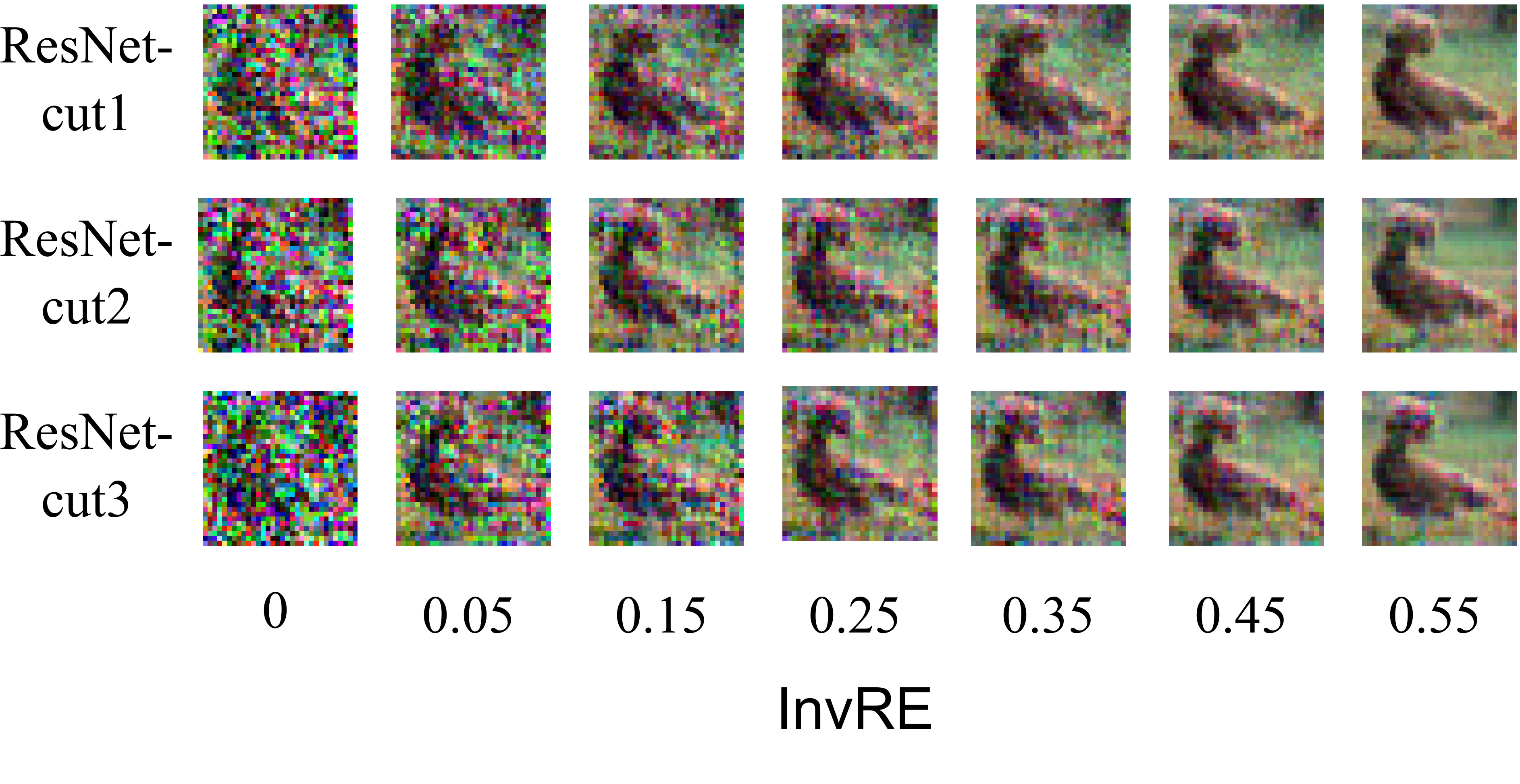} 
        \label{fig:Invre_vfl_imgs}
    \end{subfigure}
    \vspace{-7mm}
    \caption{\InvRE vs. reconstruction quality in VFL. Top: reconstructed images across increasing \InvRE levels; bottom: corresponding SSIM scores.}
    \label{fig:Invre_vfl}
\end{figure}

\subsection{Practical implications of \InvRE}
In this section, we investigate the correlation between \InvRE values and perceptual quality of reconstructed images across a wide range of \InvRE scores. 
Specifically, we divide the samples into groups based on their \InvRE values and visualize the corresponding reconstructions using representative DRAs, namely CGIR for HFL and PISTE for VFL. 
In parallel, we compute PSNR and SSIM to quantify perceptual similarity between original and reconstructed images, indicating the extent of image content leakage. Higher PSNR and SSIM denote better recovery of image details.
This combination of visual and quantitative analysis enables practitioners to directly interpret the privacy risks associated with specific \InvRE values.
Fig.~\ref{fig:Invre_hfl} and Fig.~\ref{fig:Invre_vfl} show both the reconstructed images and their corresponding SSIM scores for HFL and VFL respectively. The PSNR results with HFL and VFL settings are presented in Appendix~\ref{sec:more_results_Q4} (Fig.~\ref{fig:Invl_hfl_PSNR} and Fig.~\ref{fig:Invl_vfl_PSNR}).

As shown in Fig.~\ref{fig:Invre_hfl} and Fig.~\ref{fig:Invre_vfl}, there is a consistently strong positive correlation between \InvRE and reconstruction quality indicated by the SSIM scores in both HFL and VFL settings across different model architectures. 
\InvRE shows a strong positive correlation with SSIM, with Pearson coefficients of 0.967, 0.960, and 0.994 (with p < $10^{-4}$) for LeNet, AlexNet, and ResNet, and 0.967, 0.945, and 0.966 (with p < $10^{-3}$) for ResNet-cut1, ResNet-cut2, and ResNet-cut3, respectively. In both plots, when the \InvRE score is low (e.g., below 0.15), the reconstructed image contents are heavily degraded and lack identifiable structures. Meanwhile, the corresponding SSIM scores are also close or lower than 0.4 in both figures. According to previous works~\cite{loki, Haim2022NIPS}, SSIM lower than 0.5 usually indicates the failure of data reconstruction attacks, which is unable to recover meaningful image contents. Therefore, we can find that \InvRE score below 0.15 indicates a minimal privacy leak risk. With intermediate \InvRE values (e.g., between 0.15 and 0.45), the reconstructed images begin to reveal partial semantic features, reflecting an elevated and moderate level of privacy leakage. When the \InvRE score is high (e.g., above 0.45), the reconstructed images closely resemble the original ones, suggesting a high risk of exposing details in the images. These thresholds are supported by consistent trends in both figures across different model architectures and FL settings. Hence, \InvRE serves as a practical privacy risk auditing tool: it enables preemptive risk assessment for all samples in a training set without executing attacks.


\begin{table*}[t!]
\centering
\caption{Performance of model accuracy \small{(ACC)} and DRA error with different defenses in HFL.}
\label{table:HFL_defense}
\resizebox{0.95\linewidth}{!}{

\begin{tabular}{l|ccccl!{\vrule width 1.0pt}ccccl}
\toprule
\multirow{2}{*}{\textbf{Defense}} & \multicolumn{5}{c!{\vrule width 1.0pt}}{\textbf{CIFAR10}}                                                                & \multicolumn{5}{c}{\textbf{STL10}}                                                                \\ 
                         & $\delta$ / $\lambda$          & MSE$\uparrow$            & PSNR$\downarrow$           & SSIM$\downarrow$           & ACC$\uparrow$                    & $\delta$ / $\lambda$        & MSE$\uparrow$            & PSNR$\downarrow$           & SSIM$\downarrow$           & ACC$\uparrow$                    \\ \midrule
Clean                    & -              & 0.004          & 24.85          & 0.771          & 0.725(-)                  & -            & 0.003          & 23.43          & 0.732          & 0.865(-)                  \\ \specialrule{0em}{1.5pt}{1.5pt}
Prune                    & 0.995          & 0.024          & 16.22          & 0.410          & 0.390\small{(46.20\%)}         & 0.995         & 0.028          & 16.05          & 0.361          & 0.744\small{(13.90\%)}         \\ 
Dropout                  & 0.995          & 0.017          & 18.04          & 0.469          & 0.331\small{(54.34\%)}         & 0.970         & 0.005          & 22.70          & 0.676          & 0.424\small{(50.98\%)}         \\ \specialrule{0em}{1.5pt}{1.5pt}
\rowcolor{myyellow!40}DNP                      & 0.070          & 0.033          & 15.19          & 0.377          & 0.634\small{(12.55\%)}         & 0.050         & 0.023          & 17.32          & 0.458          & 0.813\small{(6.01\%)}          \\ 
\rowcolor{myyellow!40}InvL-DNP                 & 0.080          & \underline{0.036}          & \underline{14.63}          & \underline{0.296}          & \underline{0.642\small{(11.44\%)}}         & 0.060         & \underline{0.026}          & \underline{16.74}          & \underline{0.427}          & \underline{0.825\small{(4.62\%)}}          \\ \specialrule{0em}{1.5pt}{1.5pt}
\rowcolor{mylightgreen!80}GNP                      & 0.150          & 0.062          & 12.45          & 0.098          & 0.707\small{(2.48\%)}          & 0.030          & 0.037          & 14.58          & 0.290          & 0.831\small{(3.93\%)}          \\ 
\rowcolor{mylightgreen!80}InvL-GNP        & 0.250 & \textbf{0.065} &  \textbf{12.14} & \textbf{0.084} & \textbf{0.712\small{(1.79\%)}} & 0.070 & \textbf{0.040} & \textbf{14.10} & \textbf{0.220} & \textbf{0.848\small{(1.96\%)}} \\ \bottomrule
\end{tabular}

}
\end{table*}

\begin{table*}[t!]
\centering
\caption{Performance of model accuracy \small{(ACC)} and DRA error with different defenses in VFL.}
\label{table:VFL_defense}
\resizebox{0.95\linewidth}{!}{

\begin{tabular}{l|ccccl!{\vrule width 1.0pt}ccccl}
\toprule
\multirow{2}{*}{\textbf{Defense}} & \multicolumn{5}{c!{\vrule width 1.0pt}}{\textbf{CIFAR10}}                                                                & \multicolumn{5}{c}{\textbf{STL10}}                                                                \\ 
                         & $\delta$ / $\lambda$          & MSE$\uparrow$            & PSNR$\downarrow$           & SSIM$\downarrow$           & ACC$\uparrow$                    & $\delta$ / $\lambda$        & MSE$\uparrow$            & PSNR$\downarrow$           & SSIM$\downarrow$           & ACC$\uparrow$                    \\ \midrule
Clean                    & -              & 0.006          & 23.34          & 0.830          & 0.725(-)                  & -            & 0.009          & 21.45          & 0.732          & 0.737(-)                  \\ \specialrule{0em}{1.5pt}{1.5pt}
Prune                    & 0.900          & 0.021          & 17.76          & 0.664          & 0.518\small{(28.55\%)}         & 0.980         & 0.024          & 16.95          & 0.413          & 0.525\small{(28.76\%)}         \\ 
Dropout                  & 0.900          & 0.021          & 17.38          & 0.617          & 0.305\small{(57.93\%)}         & 0.980         & 0.023          & 17.08          & 0.398          & 0.305\small{(58.61\%)}         \\ \specialrule{0em}{1.5pt}{1.5pt}
\rowcolor{myyellow!40}DNP                      & 0.055          & 0.013          & 18.91          & 0.622          & 0.577\small{(20.41\%)}         & 0.070         & 0.022          & 16.85          & 0.402          & 0.618\small{(16.14\%)}          \\ 
\rowcolor{myyellow!40}InvL-DNP                 & 0.060          & \underline{0.014}          & \underline{18.61}          & \underline{0.605}          & \underline{0.583\small{(19.58\%)}}         & 0.080         & \underline{0.024}          & \underline{16.53}          & \underline{0.371}          & \underline{0.626\small{(15.06\%)}}          \\ \specialrule{0em}{1.5pt}{1.5pt}
\rowcolor{mylightgreen!80}GNP                      & 0.700          & 0.021          & 17.16          & 0.570          & 0.650\small{(10.34\%)}          & 0.500          & 0.039          & 14.26          & 0.350          & 0.691\small{(6.24\%)}          \\ 
\rowcolor{mylightgreen!80}InvL-GNP        & 1.000 & \textbf{0.023} & \textbf{16.43} & \textbf{0.502} & \textbf{0.684\small{(5.65\%)}} & 0.600 & \textbf{0.040} & \textbf{14.13} & \textbf{0.341} & \textbf{0.715\small{(2.98\%)}} \\ \bottomrule
\end{tabular}

}
\end{table*}

\subsection{Defense with adaptive noise perturbation}
In this section, we measure the defense performance of the proposed adaptive noise perturbation mechanism and compare it with the existing defense methods in FL systems. The empirical observations will address $\textbf{\emph{Q4}}$. 
Experiments are conducted using AlexNet and ResNet-cut2 models with CGIR and PISTE attack for HFL and VFL frameworks.
Experiment results are detailed in Table~\ref{table:HFL_defense} and Table~\ref{table:VFL_defense}.

In these tables, the ``Clean'' row represents the baseline FL system deployed without any applied defenses. The subsequent rows outline the results with various defense methods. 
For noise-based methods, defense strength is adjusted via noise variance ($\delta$); for IC-based methods, it is controlled by the pruning/dropout ratio ($\lambda$), where larger values imply stronger protection.
In all experiments, our adaptive noise is confined to the subspace spanned by more than 95\% of the total singular values. For \InvL-GNP and \InvL-ENP, the projection dimensions corresponding to the top 60\% of singular values are also set to zero.
To better illustrate the impact of each defense method on the model's utility, we report the relative drop in model accuracy (ACC) compared to the Clean baseline.

The empirical findings of this study can be summarized as follows:
\textbf{First}, compared to the noise perturbation-based defense, the IC-based defense (i.e., Prune and Dropout) offers limited protection against DRA while significantly degrading utility.
For instance, in HFL with STL10, Dropout reduces accuracy by over 50\% (from 0.865 to 0.424), yet achieves only marginal SSIM drop (0.732 to 0.676).
\textbf{Second}, \InvL-DNP consistently causes less utility degradation than DNP with comparable privacy protection strength.
In HFL on STL10, \InvL-DNP reduces ACC by 4.62\%, compared to 6.01\% for DNP. The main reason is \InvL-DNP avoids injecting noise into subspaces with small singular values, which harm the utility yet barely enhancing the protection strength (see Theorem~\ref{theorem:Bound_InvL_DNP}).
\textbf{Third}, \InvL-GNP/\InvL-ENP frameworks consistently outperform their counterparts (GNP/ ENP) and achieve the best trade-off between utility and privacy protection.
According to Table~\ref{table:VFL_defense}, the ACC drop caused by \InvL-GNP is half of that with GNP on HFL and VFL, while presenting stronger privacy protection (higher MSE, lower PSNR and SSIM). This utility-protection balance stems from the strategic noise calibration guided by Theorem~\ref{theorem:Bound_InvL_GENP}.

We also compare \InvL-DNP and \InvL-GNP/\InvL-ENP with DNP and GNP/ENP with different noise variance levels, as shown in Table~\ref{table:HFL_defense_multi_param} and Table~\ref{table:VFL_defense_multi_param} (Appendix~\ref{sec:more_results_Q5}).
The results show that the \InvL-based noise calibration methods consistently outperform the baselines across different noise variance levels, achieving equal or stronger protection with significantly lower utility loss.
In summary, the empirical findings echo the theoretical analysis behind the design of the proposed defense methods. By injecting noise only to the subspaces of the Jacobian matrix that are the most influential to the reconstruction error, \InvL-DNP and \InvL-GNP/\InvL-ENP achieve a more favorable privacy-utility balance.

\noindent\textbf{Efficiency analysis of our \InvL-based defenses.} 
The computational overhead of our \InvL-based defense framework primarily arises from two operations: the Jacobian matrix computation during backpropagation and the subsequent singular value decomposition (SVD) for spectral noise calibration. To improve efficiency, we leverage PyTorch’s \texttt{vmap} and \texttt{jacrev} for batched Jacobian computation, enabling implicit gradient aggregation and significantly reducing redundant computation and memory usage. For the SVD of Jacobian matrices $G_{\bm{x}} \in \mathbb{R}^{p \times m}$, we adopt GPU-accelerated \texttt{torch.linalg.svd()} with truncated decomposition (\texttt{full\_matrices=False}), computing only the top $d = \min(p, m)$ singular vectors. This reduces complexity to $\mathcal{O}(pmk)$ when $k \ll \min(p, m)$, while preserving key spectral components for effective noise calibration. 

\begin{table}[tb]
\centering
\caption{{Training time (s) for 100 epochs in HFL.}}
\label{table:Efficiency_HFL}
\resizebox{0.96\linewidth}{!}{
\begin{tabular}{lcccc}
\toprule
{Model}     & {Size}    & {Clean}    & {\InvL-DNP} & {\InvL-GNP} \\ \midrule
{LeNet}     & {15,826}  & {551.408}    & {1793.082}     & {735.772}       \\
{AlexNet} \ & {44,074}  & {583.687}    & {1974.017}     & {1053.765}      \\ 
{ResNet}    & {53,114}  & {1019.136}   & {3470.756}     & {2998.149}    \\ \bottomrule
\end{tabular}
}
\end{table}

\begin{table}[tb]
\centering
\caption{{Inference time (s) for 10-sample batch in VFL.}}
\label{table:Efficiency_VFL}
\resizebox{0.96\linewidth}{!}{
\begin{tabular}{lcccc}
\toprule
{Model}        &  {Size}    & {Clean}    &{\InvL-DNP} & {\InvL-ENP} \\ \midrule
{ResNet-cut1}  &  {19,392}   & {1.766e-3}     & {8.288}  & {8.287}    \\
{ResNet-cut2}  &  {208,448}  & {1.734e-3}     & {9.087}  & {9.094}   \\ 
{ResNet-cut3}  &  {503,616}  & {1.744e-3}     & {9.240}  & {9.245}   \\ \bottomrule
\end{tabular}
}
\end{table}

We present run-time measurement results for HFL and VFL in Table~\ref{table:Efficiency_HFL} and Table~\ref{table:Efficiency_VFL} respectively. In HFL, we report the total run-time cost over 100 training epochs. \InvL-based defense methods inject noise to the gradient of each training round, incurring overheads that increase linearly with the size of the model architecture. For example, ResNet has approximately four times as many parameters as LeNet. Accordingly, the training time with the \InvL-based defense methods is about three times longer than that of LeNet. In VFL, applying the defense methods during the testing time barely increases the run-time of forward inference, with the inference time consistently around 9s per batch across all model sizes.

\section{Conclusion}\label{sec:conclusion}

In this study, we propose a feasibility measure for the risk of DRAs in FL systems, called Invertibility Loss (\InvLoss), and establish its upper bound to provide a unified framework for identifying factors influencing DRA effectiveness across FL mechanisms. We extend this analysis to explain the effectiveness of existing defense mechanisms against DRAs.  
Using the theoretical upper bound, we develop \InvRE, an exact \InvLoss estimator for practical DRA risk assessment. Our results show that \InvRE consistently reflects risk levels across various FL mechanisms, model architectures, and data sources, without reliance on specific DRA methods.  
Guided by these insights, we propose two adaptive noise perturbation strategies that strategically adjust noise based on the Jacobian matrix's spectral properties. They demonstrate strong privacy protection with minimal impact on FL utility.  

\section*{Acknowledgments}
This work was supported in part by Beijing Natural Science Foundation (No. L221014), in part by National Natural Science Foundation of China (U21A20463, U22B2027, U23A20304), in part by the Systematic Major Project of China State Railway Group Co. Ltd. (No. P2023W002, No. P2024S003, and No. P2024W001-4), in part by the Science and Technology Research and Development Plan of China Railway Information Technology Group Co., Ltd. (No. WJZG-CKY-2023014 (2023A08), and No. WJZG-CKY-2024040 (2024P01)). This work was also supported by Hangzhou Qianjiang Distinguished Experts programme (2024).


\bibliographystyle{plain}
\bibliography{reference.bib}

\clearpage
\appendix

\section{Consistency Between \InvLoss and Function-Matching-Based DRA Methods}\label{sec:Consistency}
Based on Def.~\ref{def:InvL}, the process of data reconstruction attack is dedicated to deriving an approximation $\mathcal{A}_{\bm{x}}$ to the reverse transformation $\mathcal{F}^{-1}$ of $\mathcal{F}(\cdot)$, which maps back to the data instance. 
By taking $\mathcal{F}^{-1}$, the reverse transform of $\mathcal{F}$, on both sides of $\mathcal{F}(\bm{x})$ and $\mathcal{F}(\hat{\bm{x}})$ in Eq.~\eqref{eq:dra_obj} and combining $\hat{\bm{x}} = \mathcal{A}_{\bm{x}}(\mathcal{F}(\bm{x}))$ in Eq.~\eqref{eq:InvL}, we can reduce Eq.~\eqref{eq:dra_obj} to Eq.~\eqref{eq:reform_dra_obj} as below:
\begin{equation}\label{eq:reform_dra_obj}
	\begin{split}
		\hat{\bm{x}}^{*}_k &= \underset{\hat{\bm{x}} = \mathcal{A}_{\bm{x}}(\mathcal{F}(\bm{x}))}{\argmin}\,\,\|\mathcal{F}^{-1}(\mathcal{F}(\bm{x})) - \mathcal{A}_{\bm{x}}(\mathcal{F}(\bm{x}))\|^2\\
		&= \underset{\hat{\bm{x}} = \mathcal{A}_{\bm{x}}(\mathcal{F}(\bm{x}))}{\argmin}\,\,\|\mathcal{A}_{\bm{x}}(\mathcal{F}(\bm{x})) - \bm{x}\|^2.
	\end{split}
\end{equation}

\section{Missing Proofs}\label{sec:proofs}
We now provide the proofs of the theorems stated in Section~\ref{sec:method}.

\subsection{Proof of Theorem~\ref{theorem:Bound_InvL}}\label{subsec:proof_theorem1}
$\emph{Proof.}$ By the bound $\InvLoss_{\bm{x}} \leq \|(\bm{A}_{\bm{x}}\bm{G}_{\bm{x}} - \bm{I})\bm{x}\|^2 + C $ as stated in Eq.~\ref{eq:rankkattacker1},
where $\bm{A}_{\bm{x}}$ is determined by the capability of the rank-$k$ optimal DRA attacker, we proceed with the following derivation. Decompose the Jacobian matrix $\bm{G}$ using SVD, yielding $\bm{G}_{\bm{x}}=\bm{U}\bm{\Sigma}\bm{V}^{\top}$.
For a rank-$k$ optimal DRA attacker on the given instance $\bm{x}$, we have $\tilde{\bm{A}}_{\bm{x}}^{k*} = \bm{V}_{1:k}\tilde{\bm{\Sigma}}^{\dagger}\bm{U}^{\top}_{1:k}$.
Therefore,the bound on $\InvLoss_{\bm{x}}$ for the \textit{rank-$k$ optimal DRA attacker} can be expressed as:
\begin{equation}\label{eq:proof_theorem1_1}
	\begin{split}
		\InvLoss_{\bm{x}} & \leq\|(\bm{A}_{\bm{x}}\bm{G}_{\bm{x}} - \bm{I})\bm{x}\|^2 + C \\
		& \leq \sum_{k+1}^{d} (\bm{V}^{\top}\bm{x})^2 + C.
	\end{split}
\end{equation}
\qedb

\subsection{Proof of Theorem~\ref{theorem:Bound_InvL_DNP}}\label{subsec:proof_theorem2}
$\emph{Proof.}$ 
For DRA under DNP, the bound for $\InvLoss_{\bm{x}}$ can be expressed as:
\begin{equation}\label{eq:proof_theorem2_1}
	\begin{split}
		\InvLoss_{\bm{x}} & \leq \|\bm{A}_{\bm{x}}\bm{G}_{\bm{x}}(\bm{x}+\bm{\epsilon}) - \bm{x}\|^2 + C \\
		&\leq \|(\bm{A}_{\bm{x}}\bm{G}_{\bm{x}}\bm{x} - \bm{x}) + (\bm{A}_{\bm{x}}\bm{G}_{\bm{x}}\bm{\epsilon}) \|^2 + C \\
		&\leq \|\bm{A}_{\bm{x}}\bm{G}_{\bm{x}}\bm{x} - \bm{x}\|^{2} + \|\bm{A}_{\bm{x}}\bm{G}_{\bm{x}}\bm{\epsilon}\|^{2} + C,
	\end{split}
\end{equation}
where $\bm{\epsilon}$ is the noise injected into the data space under DNP.
The first term in Eq.~\eqref{eq:proof_theorem2_1} is the same as Eq.~\eqref{eq:proof_theorem1_1}.
For rank-$k$ optimal DRA attacker the second term in Eq.~\eqref{eq:proof_theorem2_1} can be expressed as:
\begin{equation}\label{eq:proof_theorem2_2}
	\|\bm{A}_{\bm{x}}\bm{G}_{\bm{x}}\bm{\epsilon}\|^{2} \leq \sum_{i=1}^{k} \frac{(\bm{V}_{i}^{\top} \bm{\epsilon})^{2}}{m}.
\end{equation}

Then we have $\InvLoss_{\bm{x}} \leq \sum_{i=k+1}^{d} (\bm{V}_{i}^{\top}\bm{x})^2 + \sum_{i=1}^{k} \frac{(\bm{V}_{i}^{\top} \bm{\epsilon})^{2}}{m} + C$.
\qedb

\subsection{Proof of Theorem~\ref{theorem:Bound_InvL_GENP}}\label{subsec:proof_theorem3}
$\emph{Proof.}$ 
For DRA under GNP/ENP, the bound for $\InvLoss_{\bm{x}}$ can be expressed as:
\begin{equation}\label{eq:proof_theorem3_1}
	\begin{split}
		\InvLoss_{\bm{x}} & \leq \|\bm{A}_{\bm{x}}(\bm{G}_{\bm{x}}\bm{x}+\bm{\epsilon}) - \bm{x}\|^2 + C \\
		&\leq \|(\bm{A}_{\bm{x}}\bm{G}_{\bm{x}}\bm{x} - \bm{x}) + (\bm{A}_{\bm{x}}\bm{\epsilon}) \|^2 + C \\
		&\leq \|\bm{A}_{\bm{x}}\bm{G}_{\bm{x}}\bm{x} - \bm{x}\|^{2} + \|\bm{A}_{\bm{x}}\bm{\epsilon}\|^{2} + C,
	\end{split}
\end{equation}
where $\bm{\epsilon}$ is the noise injected into the gradient/embedding space under GNP/ENP.
The first term in Eq.~\eqref{eq:proof_theorem3_1} is the same as Eq.~\eqref{eq:proof_theorem1_1}.
For rank-k optimal DRA attacker the second term in Eq.~\eqref{eq:proof_theorem3_1} can be expressed as:
\begin{equation}\label{eq:proof_theorem3_2}
	\|\bm{A}_{\bm{x}}\bm{\epsilon}\|^{2} \leq \sum_{i=1}^{k} \frac{(\bm{U}_{i}^{\top}\bm{\epsilon})^{2}}{\sigma _{i}^{2}p}.
\end{equation}

Then we have $\InvLoss_{\bm{x}}  \leq \sum_{i=k+1}^{d} (\bm{V}_{i}^{\top}\bm{x})^2 + \sum_{i=1}^{k} \frac{(\bm{U}_{i}^{\top}\bm{\epsilon})^{2}}{\sigma _{i}^{2}p} + C$.
\qedb

\section{Correlation Between \InvRE and Reconstruction MSE}\label{sec:E_mse}
\InvRE quantifies the average privacy risk for individual samples against attackers with varying capabilities. To maintain consistency, the reconstruction error (MSE) is similarly averaged over different attacker capabilities. Specifically, for a fixed attack algorithm, we vary the number of iteration rounds, denoted as $\text{Iters} = [\text{Iter}_{1}, \text{Iter}_{2}, \ldots, \text{Iter}_{s}]$, to simulate attackers with differing strengths.
The weighted average reconstruction error is computed by considering the probabilities associated with each attack capability. Attackers with fewer iteration rounds, representing weaker capabilities, are assigned higher probabilities, whereas attackers with more iteration rounds, representing stronger capabilities, are assigned lower probabilities. The calculation methodology is formalized as follows:
\begin{equation}\label{eq:E_mse} 
\begin{split} & E[\text{MSE}_{\bm{x}}] = \sum_{t=1}^{s} P_t\text{MSE}_{t}, \\ & \text{where} \ P_{t} = \frac{1/T_{t}}{\sum_{t=1}^{s}(1/T_{t})}, \ \text{and} \ T_{t} = \sum_{i=1}^{t} t_{i}. 
\end{split} 
\end{equation}
Here, $P_{t}$ denotes the probability of an attacker having an iteration capability of $\text{Iter}_{t}$, while $\text{MSE}_{t}$ represents the reconstruction error achieved by an attacker employing $\text{Iter}_{t}$ iterations.
When analyzing the privacy risks of different samples under a
fixed model, the \InvRE can be simplified to calculating
the inverse of the averaged \InvLoss, to visually the differences in \InvRE across samples more clearly.

\section{More Results in Experiment}\label{sec:more_results}

\begin{figure}[thb]
    \centering
    \begin{subfigure}[b]{0.45\textwidth}
        \centering
        \includegraphics[width=\textwidth]{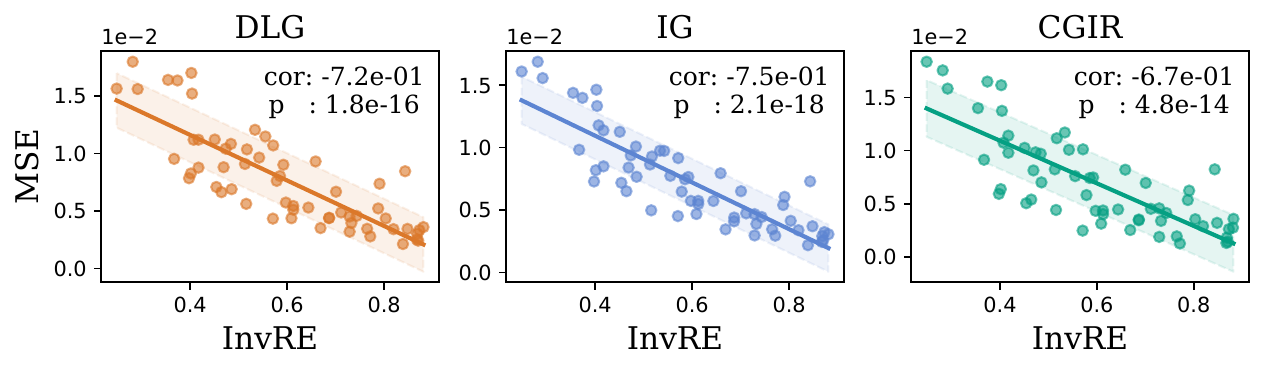} %
        \caption{STL10}
        \label{fig:sub1}
    \end{subfigure}
    \vspace{-1.5mm}
    \begin{subfigure}[b]{0.45\textwidth}
        \centering
        \includegraphics[width=\textwidth]{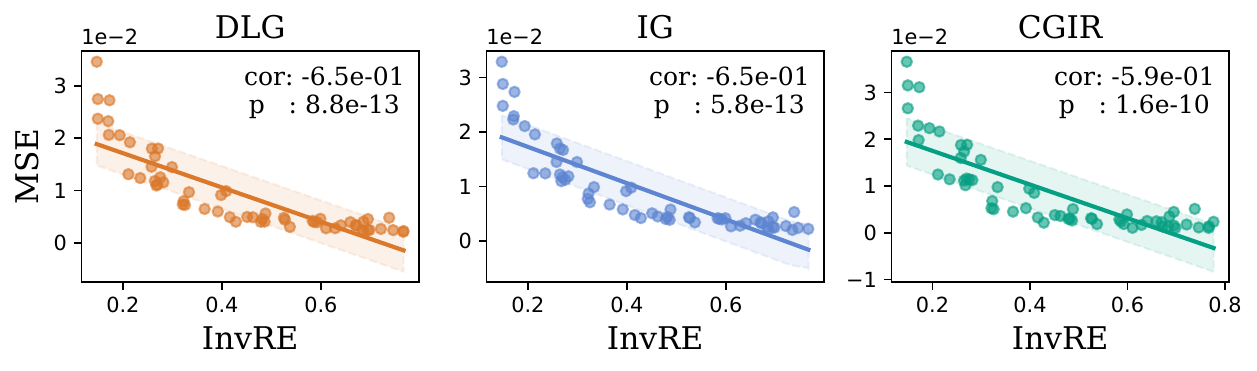} %
        \caption{SVHN}
        \label{fig:sub2}
    \end{subfigure}
    \begin{subfigure}[b]{0.45\textwidth}
        \centering
        \includegraphics[width=\textwidth]{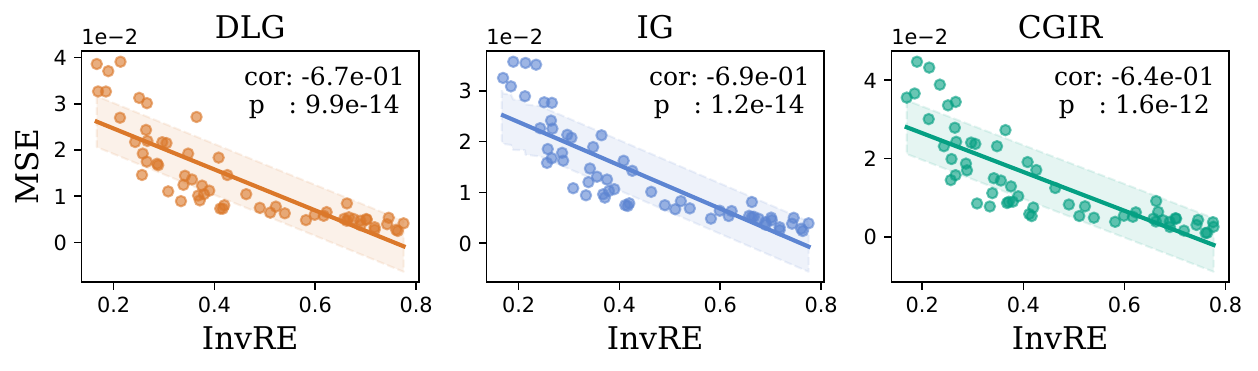} %
        \caption{ImageNet}
        \label{fig:sub3}
    \end{subfigure}
    \vspace{-3mm}
    \caption{Correlation between \InvRE and MSE for different samples trained on the LeNet model in HFL.}
    \label{fig:HFL_cross_samples}
\end{figure}

\begin{figure}[htb]
    \centering
    \begin{subfigure}[b]{0.45\textwidth}
        \centering
        \includegraphics[width=\textwidth]{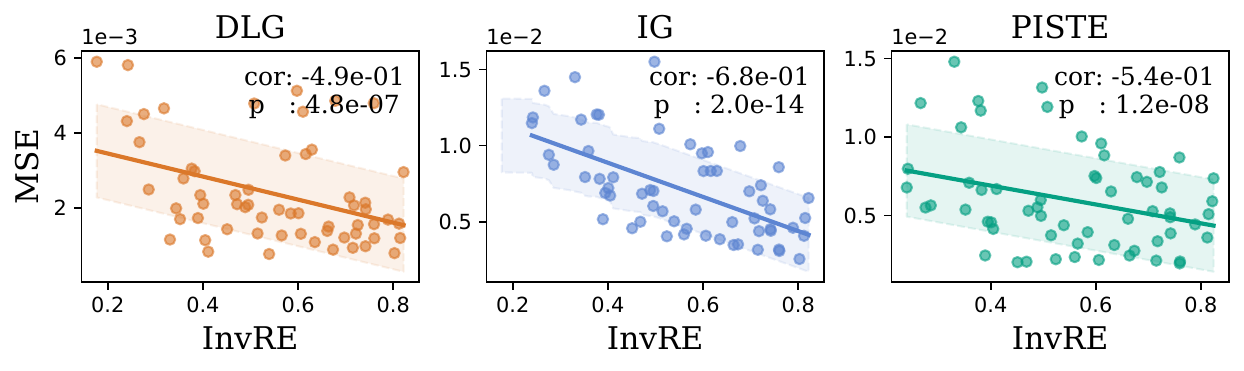} %
        \caption{STL10}
        \label{fig:vflsub1}
    \end{subfigure}
    \vspace{-1mm}
    \begin{subfigure}[b]{0.45\textwidth}
        \centering
        \includegraphics[width=\textwidth]{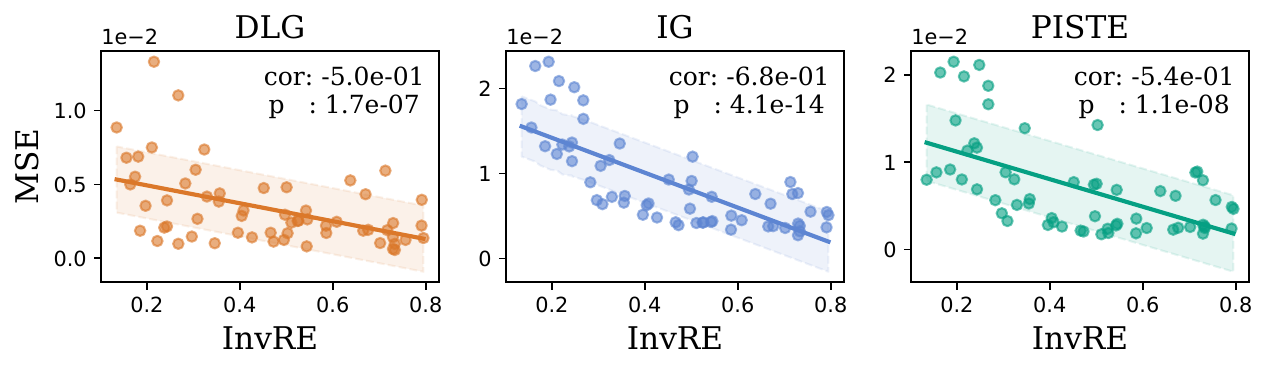} %
        \caption{SVHN}
        \label{fig:vflsub2}
    \end{subfigure}
    \begin{subfigure}[b]{0.45\textwidth}
        \centering
        \includegraphics[width=\textwidth]{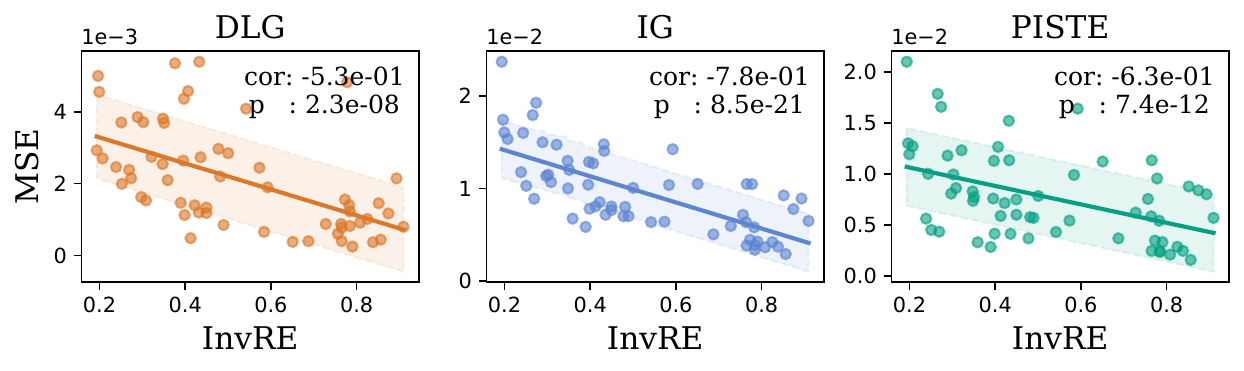} %
        \caption{ImageNet}
        \label{fig:vflsub3}
    \end{subfigure}
    \caption{Correlation between \InvRE and reconstruction MSE for different samples trained on the ResNet-cut1 in VFL.}
    \label{fig:VFL_cross_samples}
\end{figure}

\begin{figure}[b]
\centering{\includegraphics[width = 0.75\linewidth] {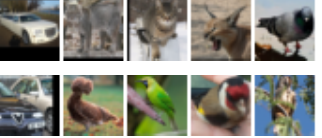}}
\vspace{1mm}
\caption{The five images with the smallest (top row) and largest (bottom row) \InvRE over STL10 on VFL system.}
\label{fig:VFL_visual_topk}
\end{figure}

\textbf{Results for validating \InvRE on input instances (\textbf{\emph{Q1})}.}
The correlation between \InvRE and reconstruction MSE using LeNet (in HFL) and ResNet-cut1 (in VFL) on the STL10, SVHN, and ImageNet datasets is presented in Fig.~\ref{fig:HFL_cross_samples} and Fig.~\ref{fig:VFL_cross_samples}. 
Additionally, Fig.~\ref{fig:VFL_visual_topk} displays the five samples with the lowest and highest \InvRE values in the VFL framework, aligning with the visualization in Fig.~\ref{fig:HFL_visual_topk}.


\begin{figure}[htb]
    \centering
    \begin{subfigure}[b]{0.485\textwidth}
        \centering
        \includegraphics[width=\textwidth]{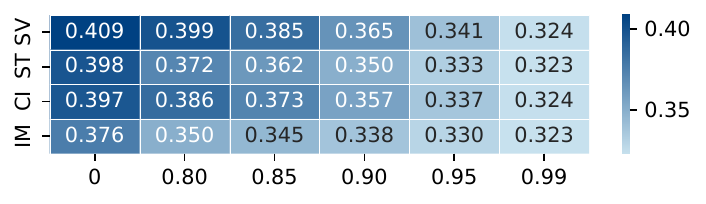} 
        \caption{Prune}
        \label{fig:hfl_prune_resnet}
    \end{subfigure}
    \begin{subfigure}[b]{0.485\textwidth}
        \centering
        \includegraphics[width=\textwidth]{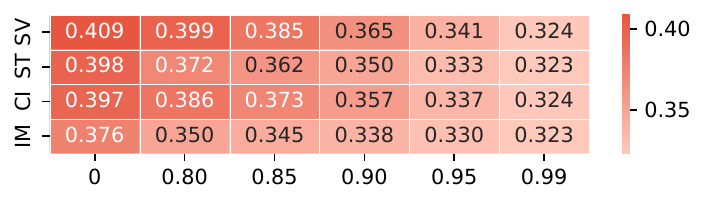} %
        \caption{Dropout}
        \label{fig:hfl_dropout_resnet}
    \end{subfigure}
    \caption{Values of \InvRE with ResNet under different defense strategies and strengths in HFL.}
    \label{fig:explain-defense-HFL-resnet}
\end{figure}

\begin{figure}[ht]
    \centering
    \begin{subfigure}[b]{0.485\textwidth}
        \centering
        \includegraphics[width=\textwidth]{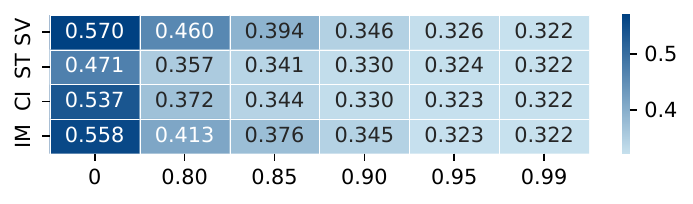} 
        \caption{Prune}
        \label{fig:vfl_prune_cut3}
    \end{subfigure}
    \begin{subfigure}[b]{0.485\textwidth}
        \centering
        \includegraphics[width=\textwidth]{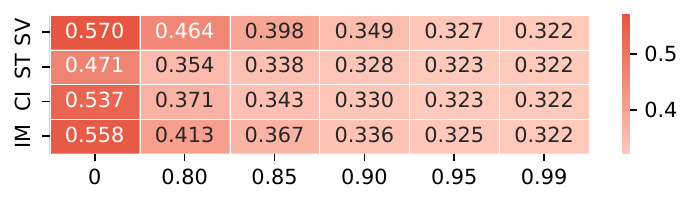} %
        \caption{Dropout}
        \label{fig:vfl_dropout_cut3}
    \end{subfigure}
    \caption{Values of \InvRE with ResNet-cut3 under different defense strategies and strengths in VFL. }
    \label{fig:explain-defense-VFL-cut3}
\end{figure}

\noindent\textbf{Results for validating existing defenses using \InvRE (\emph{Q3}).}
As shown in Theorem~\ref{theorem:Bound_InvL_DNP} and Theorem~\ref{theorem:Bound_InvL_GENP}, DNP and GNP/ENP extend Theorem~\ref{theorem:Bound_InvL} by incorporating noise terms introduced by the respective mechanisms. Increasing input noise raises the upper bound of InvLoss, reflecting enhanced defense strength.
Therefore, we additionally evaluate the \InvRE results for pruning and dropout defenses. Fig.~\ref{fig:explain-defense-HFL-resnet} and Fig.~\ref{fig:explain-defense-VFL-cut3} show \InvRE trends with varying defense strengths using ResNet (HFL) and ResNet-cut3 (VFL). The experimental results demonstrate that as the defense strength increases, the values of \InvRE consistently decrease across different datasets, aligning with the theoretical upper bound of \InvLoss presented in Sec.~\ref{subsec:explain_existing_defense}, thereby confirming its ability to explain the effectiveness of various defense mechanisms against DRAs in FL systems.


\begin{table*}[t]
\centering
\caption{{Comparison of DNP vs. \InvL-DNP and GNP vs. \InvL-GNP under various parameter settings in HFL.}}
\label{table:HFL_defense_multi_param}
\resizebox{1\linewidth}{!}{

\begin{tabular}{lcccc!{\vrule width 1.0pt}lcccc}
\toprule \specialrule{0em}{3pt}{3pt}
              & {\textbf{MSE}$\uparrow$} & {\textbf{PSNR}$\downarrow$} & {\textbf{SSIM}$\downarrow$} & {\textbf{ACC}$\uparrow$}  &               & {\textbf{MSE}$\uparrow$} & {\textbf{PSNR}$\downarrow$} & {\textbf{SSIM}$\downarrow$} & {\textbf{ACC}$\uparrow$}  \\ \specialrule{0em}{2pt}{2pt}  \midrule
{DNP-\small{0.04}}      & {0.018}        & {18.61}         & {0.52}          & {0.825} & {GNP-\small{0.03}}      & {0.037}        & {14.58}         & {0.29}          & {0.831} \\ 
{InvL-DNP-\small{0.05}} & {\underline{0.021}}        & {\underline{17.74}}         & {\underline{0.49}}          & {\textbf{0.836}} & {InvL-GNP-\small{0.07}} & {\underline{0.040}}        & {\underline{14.10}}         & {\underline{0.22}}          & {\textbf{0.848}} \\ \midrule
{DNP-\small{0.05}}      & {0.023}        & {17.32}         & {0.45}          & {0.813} & {GNP-\small{0.04}}      & {0.044}        & {13.74}         & {0.23}          & {0.825} \\ 
{InvL-DNP-\small{0.06}} & {\underline{0.026}}        & {\underline{16.74}}         & {\underline{0.42}}          & {\textbf{0.825}} & {InvL-GNP-\small{0.09}} & {\underline{0.046}}        & {\underline{13.46} }        & {\underline{0.17}}          & {\textbf{0.846}} \\ \midrule
{DNP-\small{0.06}}      & {0.026}        & {16.31}         & {0.33}          & {0.798} & {GNP-\small{0.05}}      & {0.049}       & {13.24}         & {0.19}          & {0.821}\\ 
{InvL-DNP-\small{0.07}} & {\underline{0.029}}        & {\underline{16.08}}         & {\underline{0.32}}          & {\textbf{0.802}}  & {InvL-GNP-\small{0.12}} & {\underline{0.053}}        & {\underline{12.92} }        & {\underline{0.13}}          & {\textbf{0.841}} \\ \bottomrule
\end{tabular}
}
\end{table*}

\begin{table*}[t]
\vspace{3mm}
\centering
\caption{{Comparison of DNP vs. \InvL-DNP and ENP vs. \InvL-ENP under various parameter settings in VFL.}}
\label{table:VFL_defense_multi_param}
\resizebox{1\linewidth}{!}{

\begin{tabular}{lcccc!{\vrule width 1.0pt}lcccc}
\toprule \specialrule{0em}{3pt}{3pt}
              & {\textbf{MSE}$\uparrow$} & {\textbf{PSNR}$\downarrow$} & {\textbf{SSIM}$\downarrow$} & {\textbf{ACC}$\uparrow$}  &               & {\textbf{MSE}$\uparrow$} & {\textbf{PSNR}$\downarrow$} &{ \textbf{SSIM}$\downarrow$} & {\textbf{ACC}$\uparrow$}  \\ \specialrule{0em}{2pt}{2pt}  \midrule
{DNP-\small{0.015}}      & {0.008}        & {21.82}         & {0.68}          & {0.701} & {GNP-\small{0.40}}      & {0.034}        & {14.87}         & {0.39}          & {0.699} \\ 
{InvL-DNP-\small{0.020}} & {\underline{0.009}}        & {\underline{21.60}}         & {\underline{0.66}}          & {\textbf{0.703}} & {InvL-GNP-\small{0.55}} & {\underline{0.035}}        & {\underline{14.77}}         & {\underline{0.33}}          & {\textbf{0.705}} \\ \midrule
{DNP-\small{0.065}}      & {0.020}        & {\underline{17.12}}         & {\underline{0.41}}          & {0.629} & {GNP-\small{0.50}}      & {0.039}        & {14.26}         & {0.35}          & {0.690} \\ 
{InvL-DNP-\small{0.075}} & {\underline{0.026}}        & {17.16}         & {\underline{0.41}}          & {\textbf{0.637}} & {InvL-GNP-\small{0.60}} & {\underline{0.040}}        & {\underline{14.13}}         & {\underline{0.34}}          & {\textbf{0.704}} \\ \midrule
{DNP-\small{0.07}}      & {0.022}        & {16.85}         & {0.40}          & {0.618} & {GNP-\small{0.60}}      & {0.043}        & {13.78}         & {0.32}          & {0.675}\\ 
{InvL-DNP-\small{0.08}} & {\underline{0.024}}        & {\underline{16.35}}         & {\underline{0.37}}          & {\textbf{0.626}} & {InvL-GNP-\small{0.80}} & {\underline{0.045}}        & {\underline{13.67}}         & {\underline{0.25}}          & {\textbf{0.691}} \\ \bottomrule
\end{tabular}

}
\end{table*}

\begin{figure}[t]
\centering{\includegraphics[width = 0.85\linewidth] {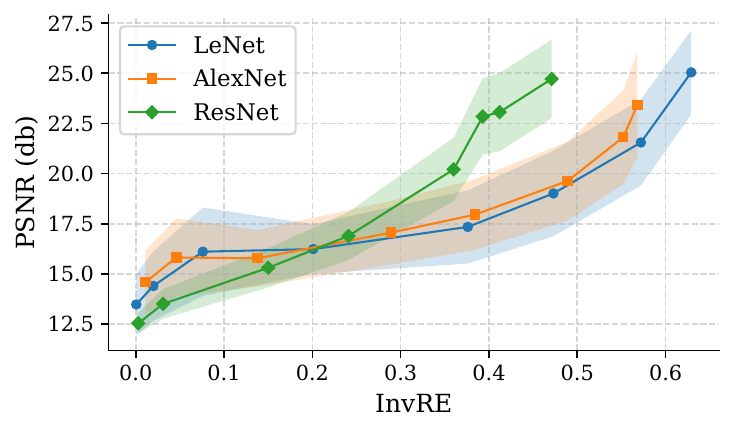}}
\caption{{\InvRE vs. PSNR (higher means successful reconstruction) in HFL.}}
\label{fig:Invl_hfl_PSNR}
\end{figure}

\begin{figure}[t]
\centering{\includegraphics[width = 0.85\linewidth] {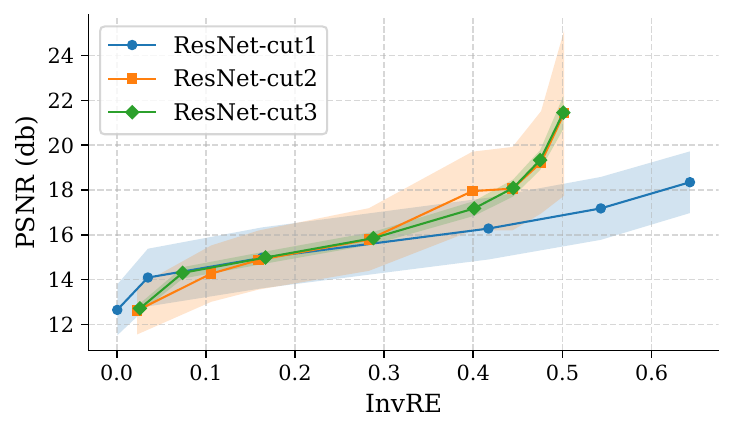}}
\caption{{\InvRE vs. PSNR (higher means successful reconstruction) in VFL.}}
\label{fig:Invl_vfl_PSNR}
\end{figure}

\noindent\textbf{Results of practical implications of \InvRE (\emph{Q4}).}\label{sec:more_results_Q4}
To complement the SSIM-based analysis presented in the main text, we provide additional results with the metric of PSNR in Fig.~\ref{fig:Invl_hfl_PSNR} and Fig.~\ref{fig:Invl_vfl_PSNR} for HFL and VFL, respectively.
In the HFL setting (Fig.~\ref{fig:Invl_hfl_PSNR}), the Pearson correlation coefficients between the reported \InvRE and the PSNR scores are consistently strong, presenting 0.936, 0.940, and 0.983 for LeNet, AlexNet, and ResNet respectively with the p-values less than $2e^{-5}$.  In the VFL setting (Fig.~\ref{fig:Invl_vfl_PSNR}), for ResNet-cut1, ResNet-cut2, and ResNet-cut3, the Pearson correlation coefficients are 0.976, 0.962, and 0.945 with corresponding p-values less than $4e^{-4}$ respectively. \InvRE shows consistently strong positive correlation with the PSNR scores, indicating that \InvRE increases/decreases proportionally as the PSNR scores.

\vspace{3mm}

When \InvRE is lower than 0.15, PSNR drops below 17dB in HFL and 15dB in VFL, indicating severe reconstruction loss per prior studies~\cite{CGIR,Sara2019ImageQA}. As shown in Fig.~\ref{fig:Invre_hfl} and Fig.~\ref{fig:Invre_vfl}, reconstructions in this range lack clear structure, signaling failed attacks. As \InvRE rises from 0.15 to 0.4, PSNR improves to over 22.5dB (HFL) and 18dB (VFL), revealing more recognizable textures with moderate quality. When \InvRE exceeds 0.4, PSNR reaches 25dB (HFL) and 22dB (VFL), aligning with prior definitions~\cite{IG, shi2024scalemiascalablemodelinversion} of successful reconstruction. Corresponding images show clear textures of the originals.
In summary, higher \InvRE values are closely associated with higher PSNR or SSIM scores, indicating more accurate reconstructions of image details. This consistent trend across multiple image reconstruction metrics reinforces the reliability of \InvRE as a proxy for privacy leakage.

\noindent\textbf{Results for comparing with other methods (\emph{Q5}).}\label{sec:more_results_Q5}
Table~\ref{table:HFL_defense_multi_param} and Table~\ref{table:VFL_defense_multi_param} compare \InvL-DNP, and \InvL-GNP/\InvL-ENP with their respective baselines (DNP and GNP/ENP) across a range of parameter configurations on STL10 for HFL and VFL, respectively. 
As the results show, our \InvL-based defense methods consistently achieve better task performance across all settings, while maintaining comparable or stronger privacy protection.
These findings further validate the effectiveness of our spectrally calibrated noise injection strategy in achieving a better privacy-utility trade-off across diverse conditions.

\section{Additional Experiments on LOKI~\cite{loki}}\label{sec:more_results_loki}
In this section, we evaluate our \InvL-based defense against LOKI~\cite{loki}, which steals client data via a malicious HFL server by modifying the model architecture.
Specifically, LOKI inserts an attack module at the beginning of the original model, consisting of a convolutional layer followed by two fully connected (FC) layers.  
Image leakage occurs through gradients computed with respect to the weights connecting the convolutional output to the first FC layer.  
We use ResNet as the base model and STL10 as the dataset, with 20 neurons in the FC layer and a batch size of 10. \InvRE scores are evaluated on the first 100 STL10 samples. Results are shown in Table~\ref{table:defense_loki}.
As shown, our \InvL-DNP and \InvL-ENP achieve a more optimal balance between privacy protection and model utility compared to existing noise perturbation defense methods.

\vspace{2mm}
\begin{table}[th]
\centering
\caption{Performance of model accuracy (ACC) and DRA error by LOKI under different defenses.}
\label{table:defense_loki}
\resizebox{0.98\linewidth}{!}{
\begin{tabular}{lcccccc}
\toprule  
         & {$\delta$} & {MSE$\uparrow$} & {PSNR$\downarrow$} & {SSIM$\downarrow$} & {ACC$\uparrow$} \\  \midrule
{DNP}      & {0.03}  & {0.0024}    &    {26.47}  &   {0.905}   &   {0.833}  \\ 
{\InvL-DNP} & {0.07}  &  {\textbf{0.012}}   &   {\textbf{24.94}}   &  {\textbf{0.772}}    & {\textbf{0.840}}    \\ \midrule
{GNP}      & {0.001}  &  {0.060}   &   {12.35}   &   {0.396}   &  {0.802}    \\ 
{\InvL-GNP} & {0.005}  &  {\textbf{0.081}}   &    {\textbf{11.09}}  &   {\textbf{0.373}}   & {\textbf{0.813}}    \\ \bottomrule
\end{tabular}
}
\end{table}

\end{document}